\documentclass{article}

\usepackage{microtype}
\usepackage{graphicx}
\usepackage{booktabs} 

\usepackage{hyperref}

\usepackage[accepted]{icml2024}

\usepackage{amsmath}
\usepackage{amssymb}
\usepackage{mathtools}
\usepackage{amsthm}

\usepackage[capitalize,noabbrev]{cleveref}

\theoremstyle{plain}

\theoremstyle{definition}

\theoremstyle{remark}

\usepackage[textsize=tiny]{todonotes}

\usepackage{microtype}
\usepackage{booktabs} 
\usepackage{multirow}
\usepackage{amsmath}
\usepackage{amsthm}
\usepackage{amsfonts}
\usepackage{times}
\usepackage{soul}
\usepackage{url}
\usepackage[utf8]{inputenc}
\usepackage[small]{caption}
\usepackage{booktabs}
\usepackage{graphicx}
\usepackage{subcaption}

\def\onedot{.}
\def\eg{\emph{e.g}\onedot} 
\def\ie{\emph{i.e}\onedot}

\usepackage{xcolor}

\newcommand{\specialcell}[2][c]{%
	\begin{tabular}[#1]{@{}c@{}}#2\end{tabular}
}

\usepackage{makecell, array, booktabs, multirow, caption}

\usepackage{adjustbox}
\usepackage{wrapfig}

\icmltitlerunning{Probabilistic Subgoal Representations for Hierarchical Reinforcement Learning}

\begin{document}

\twocolumn[
\icmltitle{Probabilistic Subgoal Representations for Hierarchical Reinforcement learning}



\icmlsetsymbol{equal}{*}



\begin{icmlauthorlist}
\icmlauthor{Vivienne Huiling Wang}{equal,xxx}
\icmlauthor{Tinghuai Wang}{equal,comp}
\icmlauthor{Wenyan Yang}{yyy}
\icmlauthor{Joni-Kristian K\"{a}m\"{a}r\"{a}inen}{yyy}
\icmlauthor{Joni Pajarinen}{xxx}
\end{icmlauthorlist}

\icmlaffiliation{xxx}{Department of Electrical Engineering and Automation, Aalto University, Finland}
\icmlaffiliation{comp}{Huawei Helsinki Research Center, Finland}
\icmlaffiliation{yyy}{Computing Sciences, Tampere University, Finland}

\icmlcorrespondingauthor{Vivienne Huiling Wang}{vivienne.wang@aalto.fi}

\icmlkeywords{Hierarchical Reinforcement Learning}

\vskip 0.3in
]



\printAffiliationsAndNotice{\icmlEqualContribution} 

\begin{abstract}

    In goal-conditioned hierarchical reinforcement learning (HRL), a high-level policy specifies a subgoal for the low-level policy to reach. Effective HRL hinges on a suitable subgoal representation function, abstracting state space into latent subgoal space and inducing varied low-level behaviors. Existing methods adopt a subgoal representation that provides a deterministic mapping from state space to latent subgoal space. Instead, this paper utilizes Gaussian Processes (GPs) for the first probabilistic subgoal representation. Our method employs a GP prior on the latent subgoal space to learn a posterior distribution over the subgoal representation functions while exploiting the long-range correlation in the state space through learnable kernels. This enables an adaptive memory that integrates long-range subgoal information from prior planning steps allowing to cope with stochastic uncertainties. Furthermore, we propose a novel learning objective to facilitate the simultaneous learning of probabilistic subgoal representations and policies within a unified framework. In experiments, our approach outperforms state-of-the-art baselines in standard benchmarks but also in environments with stochastic elements and under diverse reward conditions. Additionally, our model shows promising capabilities in transferring low-level policies across different tasks.

    \end{abstract}
    
    \section{Introduction}

     Addressing intricate issues that require long-term credit assignment remains a significant hurdle in reinforcement learning (RL). Hierarchical deep reinforcement learning (HRL) stands out in this context, demonstrating potential  in handling various temporally extended tasks with sparse rewards. This is achieved through its ability to facilitate control at multiple temporal scales using a hierarchical framework.  Goal-conditioned HRL methods, in which the higher-level policies periodically set subgoals for lower-level policies and the lower level is intrinsically rewarded for reaching those subgoals, have long held the promise to be an effective paradigm in HRL \citep{DayanH92,schmidhuber1993planning,kulkarni2016hierarchical,vezhnevets2017feudal,NachumGLL18,LevyKPS19,ZhangG0H020,li2020learning,LiZWYZ22,wang2023state}. 
        
        The subgoal representation function in goal-conditioned HRL maps the state space to a latent subgoal space. Learning an appropriate subgoal representation function is critical to the performance and stability of goal-conditioned HRL. Since the subgoal space corresponds to the high-level action space, the subgoal representation contributes to the stationarity of the high-level transition functions. Furthermore, the low-level reward function, \ie, intrinsic rewards, is defined in latent subgoal space in goal-conditioned HRL, and low-level behaviors can be induced by dynamically changing subgoal space as well. As such, a proper abstract subgoal space contributes to the stationarity of hierarchical policy learning.

     \begin{figure*}[!t]
        \centering
        \begin{adjustbox}{valign=t}
        \begin{minipage}{0.45\textwidth}
        \centering
            \includegraphics[width=0.99\linewidth,height=6.1cm]{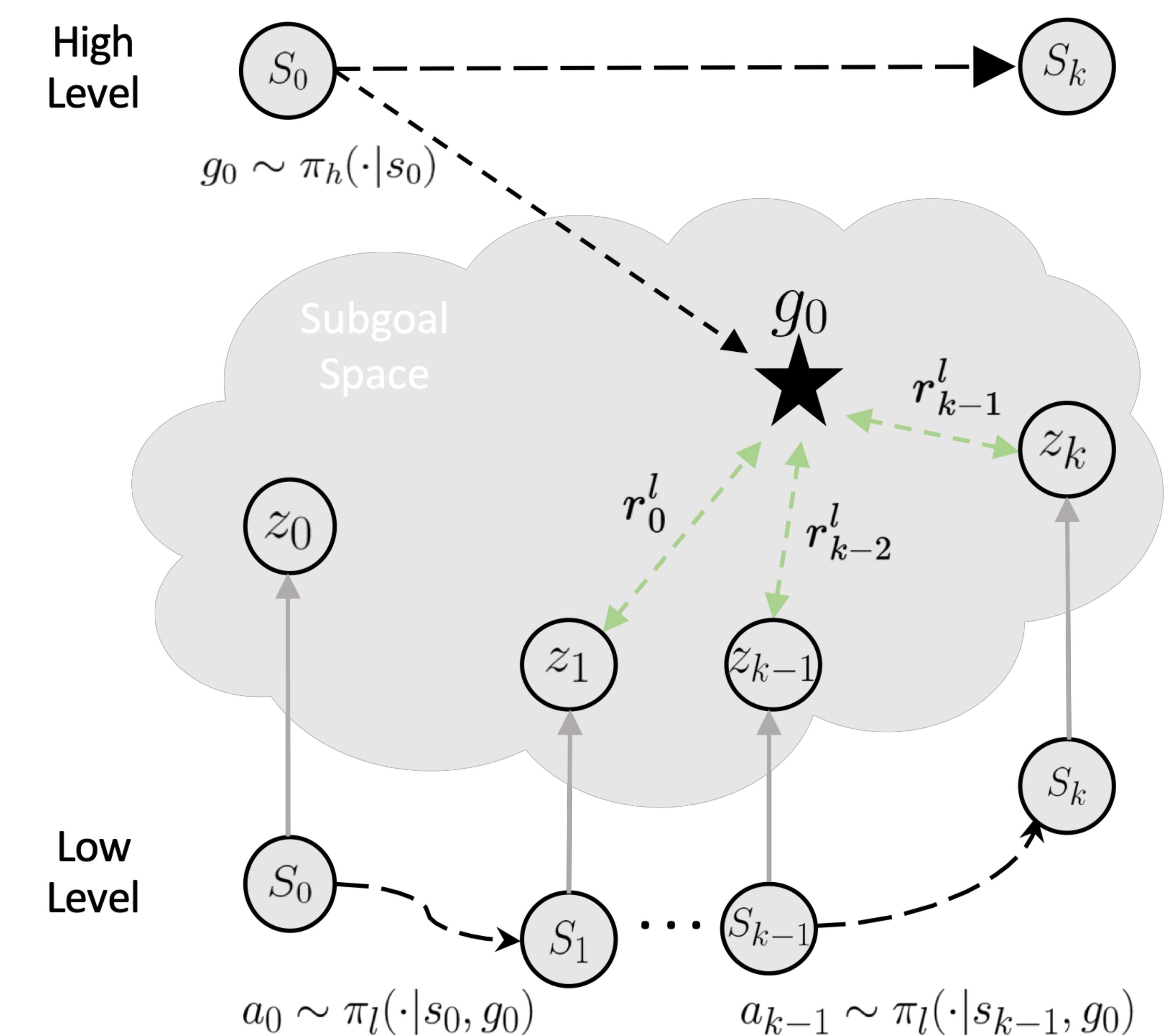}
            \caption{A schematic illustration of the hierarchical policy execution. One high-level step corresponds to k low-level steps. The negative Euclidean distance in the latent space provides intrinsic rewards for the low-level policy.}
            \label{fig:diag}
        \end{minipage}
        \end{adjustbox}
        \hfill 
        \begin{adjustbox}{valign=t}
        \begin{minipage}{0.45\textwidth}
        \centering
            \includegraphics[width=0.7\linewidth,height=5.6cm]{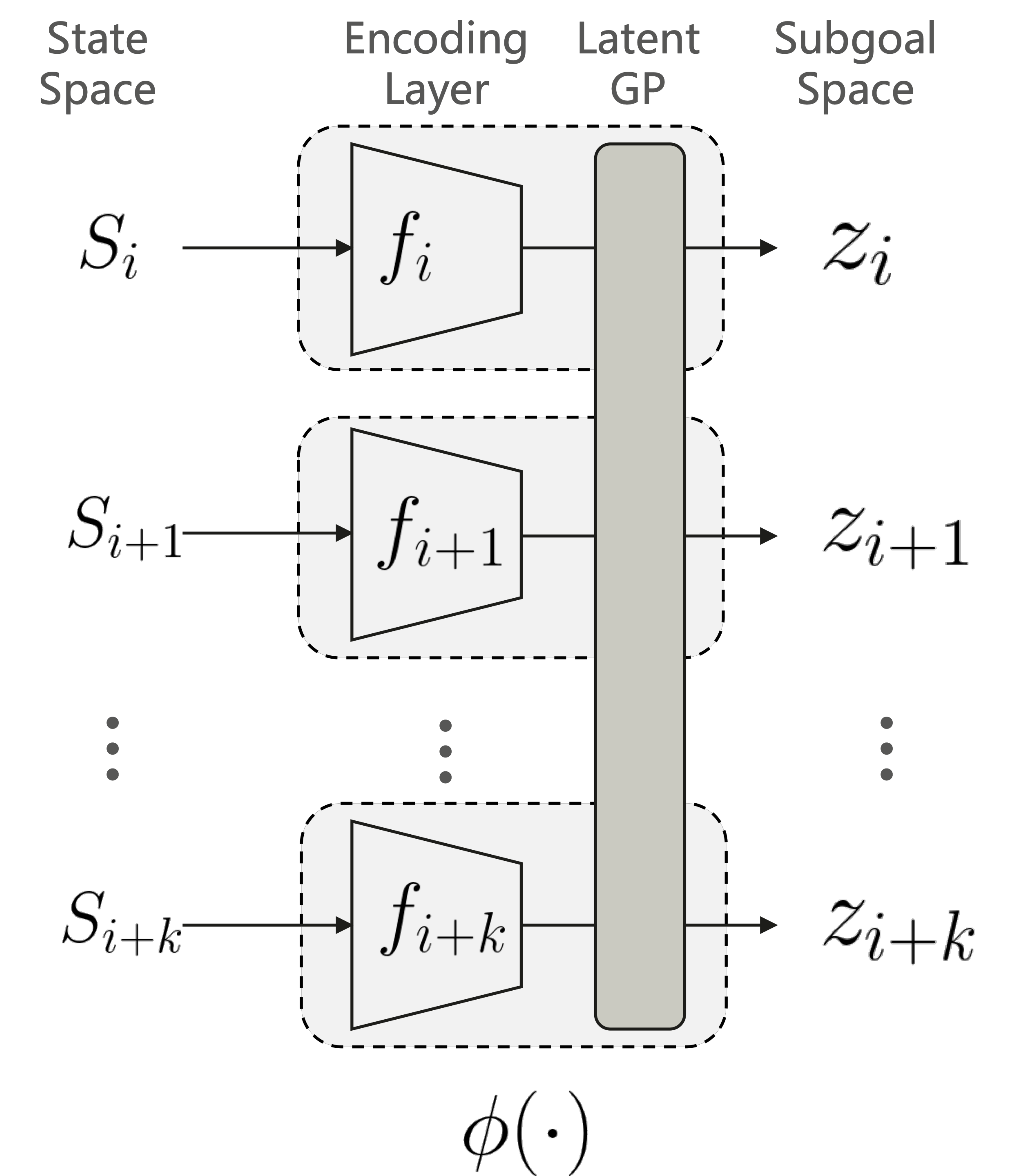}
            \caption{The representation function consists of an encoding layer and a latent GP layer. Taking as input the state $\mathbf{s}$, the encoding layer comprises a neural network to generate an intermediate latent space representation $\mathbf{f}$, which will be transformed by the GP layer to produce the final subgoal representation $\mathbf{z}$.}
            \label{fig:network}
        \end{minipage}
        \end{adjustbox}
    \end{figure*}

        A wide variety of subgoal representations have been investigated, ranging from directly utilizing the state space \citep{LevyKPS19} or hand-crafted space \citep{NachumGLL18}, to end-to-end learning without explicit objectives \citep{vezhnevets2017feudal} or deterministic representations learned by imposing local constraints \citep{li2020learning}. Notably, previous works utilizing deterministic subgoal representation functions, \eg, \citet{li2020learning},  often struggle to adapt to unforeseen or novel states, as they rely on fixed representations that may not accurately capture the variability and unpredictability of dynamic environments. The deterministic mapping inherently lacks the capacity to account for stochastic uncertainties, some arising directly from environmental stochasticity, while others can be attributed to unexplored areas of the state space. This potentially limits the exploration capacity of hierarchical policies and leading to convergence on suboptimal solutions. In scenarios involving novel state regions, deterministic mapping may not have sufficient historical data to form an appropriate subgoal representation, possibly underfitting the learning objective and failing to accurately capture the dynamic nature of new state regions.  This limitation often results in poor performance and hinders the agent's goal achievement.  Although the active exploration strategy proposed by \citet{LiZWYZ22} aims to mitigate some of these issues in \citet{li2020learning}, the fundamental limitations associated with deterministic mapping and the emphasis on short-term smoothness — stemming from local constraints — continue to restrict effective exploration and impact the stationarity of training hierarchical policies.
    
    To address these limitations, we propose a novel GP based approach to learning probabilistic subgoal representations for goal-conditioned HRL (HLPS). HLPS considers a continuum of possible subgoal representation functions rather than a single, predetermined mapping. This probabilistic formulation which better reflects the uncertain nature of the environment, is achieved by harnessing a nonparametric GP prior on the latent subgoal space to learn a posterior distribution over the subgoal representation functions. This probabilistic nature of HLPS enables the adaptation to new observations without the need for redefining the model structure. As more state regions are explored, the posterior distribution of subgoal representation functions updates, refining the model's predictions and making it more robust to unseen states. With the learnable kernels, HLPS can adaptively exploit the long-range correlations in the state space which allows to capture and utilize the underlying patterns and relationships within the state space. We further propose a novel learning objective that cohesively integrates the learning of probabilistic subgoal representations and hierarchical policies. Leveraging the nature of Markov chains, we also present a lightweight subgoal representation formulation for online inference, harnessing the state-space form GP, which efficiently fuses the subgoal information from an arbitrary number of previous planning steps with a constant computational and memory complexity.


        We benchmark our method on challenging continuous control tasks in both deterministic and stochastic settings with dense or sparse external rewards. Experimental results empirically demonstrate that our method is capable of generating stable subgoal representations which, on the one hand, contribute to the stationarity in both the high-level state transition and the low-level reward functions and, on the other hand, facilitates transferable low-level policies between tasks. The advantages of this first probabilistic subgoal representation within HRL manifest as increased sample efficiency, heightened resilience against stochastic uncertainties, and a marked improvement in asymptotic performance when benchmarked against leading HRL methods.

    Our main contributions are as follows:
    \begin{itemize}
        \item We introduce \textbf{probabilistic subgoal representations} for goal-conditioned HRL, which consider a continuum of possible subgoal representation functions rather than a single, deterministic mapping. This allows us to better reflect the stochastic uncertainties of the environment and to adapt to new observations without redefining the model structure.
        \item We propose a \textbf{novel learning objective} that cohesively integrates the learning of probabilistic subgoal representations and hierarchical policies. 
        \item We present a \textbf{lightweight subgoal representation formulation} for online inference, harnessing the state-space form GP, which efficiently fuses the subgoal information from an arbitrary number of previous planning steps with a constant computational and memory complexity.
    \end{itemize}

        \section{Preliminaries}
        
        The interaction between the agent and environment is generally modeled as a Markov Decision Process (MDP). Consider a goal-conditioned MDP which is represented by a tuple:  $\text{MDP}=<\mathcal{S}, \mathcal{G},  \mathcal{A}, \mathcal{P}, \mathcal{R}, \gamma>$, where $\mathcal{S}$ is a state space, 
        $\mathcal{G}$ is the subgoal set, $\mathcal{A}$ is an action set, $\mathcal{P}: \mathcal{S} \times \mathcal{A} \times \mathcal{S} \to [0, 1]$ is a state transition function, $\mathcal{R}: \mathcal{S} \times \mathcal{A} \to \mathbb{R}$ is a reward function, and $\gamma \in [0, 1)$ is a discount factor. We consider an HRL framework with two levels following \citet{NachumGLL18} as illustrated in Fig. \ref{fig:diag}: the high-level policy $\pi_h(g|s)$ which operates at a coarser layer and generates a high-level action, \ie, subgoal, and the low-level policy $\pi_l(a|s, g)$ which aims to achieve these subgoals. The high-level policy maximizes external reward by generating subgoals, \ie, $g_i \sim \pi_h (\cdot| s_i) \in \mathcal{G}$, every $k$ timesteps when $i\equiv 0~($mod $k)$. The low-level policy maximizes intrinsic reward associated with the subgoals by executing the primitive action.

        \section{Method}
        
        In this section, we present our  Gaussian Process based probabilistic subgoal representation. Firstly, we introduce a two-level goal-conditioned HRL framework with state-kernel GP prior. Then we present GP latent-state batch estimation and training objective, which is followed by a lightweight online planning scheme.

        \subsection{Framework}
        
        We define the subgoal $g$ in the two-level HRL framework introduced by  \citet{NachumGLL18} in a low dimensional space abstracted by representation function $\phi(\mathbf{s}): \mathbf{s} \mapsto \mathbb{R}^d$. Our method learns $\phi(\mathbf{s})$ simultaneously with the hierarchical policy. Specifically, we train the low-level policy $\pi_l(a|s, g)$ with an intrinsic reward function defined as the negative Euclidean distance in the latent subgoal space, \ie, $r_l (s_i, a_i, s_{i+1}, g_i) = - ||\phi(s_{i+1}) -g_i||_2$. The high-level policy is trained to maximize the extrinsic reward $r_{i}^h$ defined as $r_{i}^h=\sum _{t=i}^{i+k-1} r_t^{\text{env}}, i=0,1,2,\cdots$,
        where $r_t^{\text{env}}$ is the reward from the environment.  Our framework adopts the off-policy algorithm SAC \citep{haarnoja2018soft} for each level in the HRL structure, which generalizes the standard RL objective by augmenting it with an entropy term, \ie,  
    $\pi^* = \underset{\pi}{\mathrm{argmax}} \sum_t \mathbb{E}_{(s_i,a_i) \sim \rho_\pi} \left[ r(s_i, a_i) + \alpha \mathcal{H}(\pi(\cdot | s_i)) \right].$
     Nonetheless, it is important to note that our method is agnostic to the specific HRL framework used. 
     
     As illustrated in Fig. \ref{fig:diag}, the high-level policy samples a subgoal $\mathbf{g}$ from the latent subgoal space abstracted by the representation function $\phi(\mathbf{s})$. The low-level controller then strives to achieve these subgoals, with $\mathbf{z}=\phi(\mathbf{s})$ serving as the mapping from the state space to the subgoal space. As shown in Fig. \ref{fig:network}, the representation function $\phi(\mathbf{s})$ consists of an encoding layer and a latent GP layer. The encoding layer comprises a neural network to generate an intermediate latent space representation $\mathbf{f}$ by taking as input the state $\mathbf{s}$, which will be transformed by the GP layer to produce the final subgoal representation $\mathbf{z}$.  
        
        \subsection{Probabilistic Subgoal Representation}
        In order to specify a complete probabilistic model connecting state and subgoal spaces, a prior distribution for the latent subgoal $\mathbf{z}$ has to be defined.  To this end, we impose GP priors to all $\mathbf{z}$ to model the stochastic uncertainties in subgoal space. Some of these uncertainties arise directly from environmental stochasticity, while others may be attributed to the unexplored regions of the state space. Specifically, we model the intermediate  latent space representation $\mathbf{f}$ as a noise-corrupted version of the true latent subgoal space representation $\mathbf{z}$, and the inference can be stated as the following GP regression model:
        \begin{equation}
        \begin{array}{l}
        \mathbf{z}_i \sim \mathcal{G P}\left(0, \kappa\left(\mathbf{s}_i, \mathbf{s}_j \right)  \right),   \\
        \mathbf{f}_i =\mathbf{z}_i +\epsilon, \epsilon \sim \mathcal{N}(0, \sigma),   
        \label{eq:gp}
        \end{array}
        \end{equation}
        where the noise variance $\sigma^2$ is a learnable parameter of the likelihood model, and $\kappa\left(\mathbf{s}_i, \mathbf{s}_j\right) $ is a positive-definite kernel function. 
        
        By modeling the uncertainties in the subgoal space with GP priors, the mapping from state space to subgoal space transforms from a deterministic function into a probabilistic distribution of subgoal representation functions.  GP priors also define a probabilistic prior on the intermediate latent space which encodes for \emph{a priori} knowledge that similar states should be mapped to more relevant latent subgoal representations than those mapped from distinct states.  Such prior knowledge could be encoded by the kernel function, \ie, $\kappa\left(\mathbf{s}_i, \mathbf{s}_j \right) $, defined over a distance in state space. Our insight is that the long-range correlation in the state space could be exploited through learnable kernel function. We define the prior to be mean square continuous, once differentiable, and stationary in state space for the latent space processes \citep{williams2006gaussian}. Since the latent functions are intended to model the intrinsic structure of the state space, the latent  space is expected to behave in a smooth and continuous fashion which is satisfied by Mat\'{e}rn kernel \citep{williams2006gaussian}, 
        \begin{equation}
        \kappa\left(\mathbf{s}_i, \mathbf{s}_j \right)=\gamma^{2}\left(1+\frac{\sqrt{3} D\left(\mathbf{s}_i, \mathbf{s}_j \right)}{\ell}\right) \exp \left(-\frac{\sqrt{3} D\left(\mathbf{s}_i, \mathbf{s}_j \right)}{\ell}\right).
        \end{equation}
        This kernel encodes the similarity between two states $\mathbf{s}_i$ and $\mathbf{s}_j$ in latent subgoal space subject to the distance function $D(\cdot)$ which is defined as $\ell^2$-norm. The learnable hyperparameters $\gamma^{2}$ and $\ell$ characterize the magnitude and length-scale of the processes respectively.

        The inference problem in Eq. \ref{eq:gp} can be solved for an unordered set of states, and the posterior mean and covariance are given by  \cite{williams2006gaussian}:
        \begin{equation}
        \begin{array}{l}
        \mathbb{E}[\mathbf{Z} \mid \mathbf{S}, \mathbf{F}]=\mathbf{C}\left(\mathbf{C}+\sigma^{2} \mathbf{I}\right)^{-1} \mathbf{F},\\
        \mathbb{V}[\mathbf{Z} \mid \mathbf{S}, \mathbf{F}]=\operatorname{diag}\left(\mathbf{C}-\mathbf{C}\left(\mathbf{C}+\sigma^{2} \mathbf{I}\right)^{-1} \mathbf{C}\right),
        \label{eq:batch}
            \end{array}
        \end{equation}
        where $\mathbf{Z} = \left( \mathbf{z}_1, ~ \mathbf{z}_2, ~ \cdots, ~ \mathbf{z}_N  \right)$ are the set of subgoal representations,  $\mathbf{F} = \left( \mathbf{f}_1, ~ \mathbf{f}_2, ~ \cdots, ~ \mathbf{f}_N  \right)$ are the set of intermediate latent representations from encoding layer, and $\mathbf{C}_{i,j} = \kappa\left(\mathbf{s_i} , \mathbf{s_j} \right)$ represents the covariance matrix. The true latent space representation, \ie, the subgoal representation, $\mathbf{Z}$ can be restored by taking the posterior mean of the GP.  	
        
        \subsection{Learning Objective}
    
        In order to learn the hyperparameters of our probabilistic subgoal representation, \ie,  $\sigma^2$, $\gamma^{2}$ and $\ell$, we propose a learning objective as follows:
        \begin{equation}
        \mathcal{L} = \frac{\Delta_\mathbf{f}^{1}}{\Delta_\mathbf{f}^{k}}\text{log}(1+\text{exp}(\Delta_\mathbf{z}^{1}-\Delta_\mathbf{z}^{k})), 
        \label{eq:objective}
        \end{equation}
        where $\Delta_\mathbf{f}^{1} \propto ||\mathbf{f}_{i}-\mathbf{f}_{i+1}||$, $\Delta_\mathbf{f}^{k} \propto ||\mathbf{f}_{i}-\mathbf{f}_{i+k}||$, $\Delta_\mathbf{z}^{1} \propto ||\mathbf{z}_{i}-\mathbf{z}_{i+1}||$ and $\Delta_\mathbf{z}^{k} \propto ||\mathbf{z}_{i}-\mathbf{z}_{i+k}||$. The logarithmic term in our proposed objective is designed to minimize the distance between low-level state transitions (\(\Delta_{\mathbf{z}}^{1}\)) in the latent subgoal space, while maximizing the distance for high-level state transitions (\(\Delta_{\mathbf{z}}^{k}\)). We employ the softplus function \citep{dugas2000incorporating} over the hinge loss for two main reasons. Firstly, it eliminates the need for a margin hyperparameter, thus simplifying the optimization process. Secondly, the softplus function provides continuous gradients, as opposed to the discontinuous gradients around margin planes seen in the hinge loss, facilitating finer adjustments within the subgoal space \(\mathbf{Z}\). Furthermore, to enhance feature discrimination and the interaction between \(\mathbf{F}\) and \(\mathbf{Z}\), we use the ratio \(\frac{\Delta_{\mathbf{f}}^{1}}{\Delta_{\mathbf{f}}^{k}}\) as a relative distance measure in \(\mathbf{F}\) for the auxiliary loss. This approach promotes closer intermediate latent representations for low-level state transitions with smaller ratios and greater separation for high-level transitions with larger ratios, focusing on the relative ratio rather than the absolute difference.

        This objective is specifically designed for modeling the stochasticity of subgoal space (ratio term) while facilitating smooth and yet discriminative subgoal representation learning in GP latent space (logarithm term). Rather than learning a deterministic mapping from state space to subgoal space, our probabilistic approach explicitly represents subgoals at a finite number of support points, \ie, $\mathbf{S} = \{ \mathbf{s}_i, ~ \mathbf{s}_{i+1}, ~ \mathbf{s}_{i+k} \}$, and let the GPs generalize to the entire space through the kernel function with learned hyperparameters.

    \begin{figure*}[!t]
        \centering
        \begin{subfigure}{0.16\linewidth}
            \captionsetup{skip=0pt, position=below}
            \includegraphics[width=\linewidth]{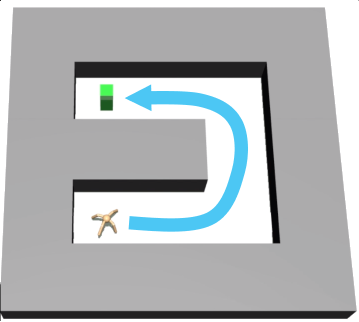}
        \end{subfigure}
        \begin{subfigure}{0.16\linewidth}
            \captionsetup{skip=0pt, position=below}
            \includegraphics[width=\linewidth]{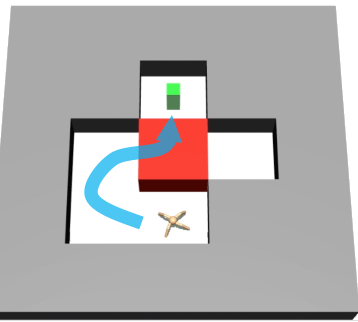}
        \end{subfigure}
        \begin{subfigure}{0.16\linewidth}
            \captionsetup{skip=0pt, position=below}
            \includegraphics[width=\linewidth]{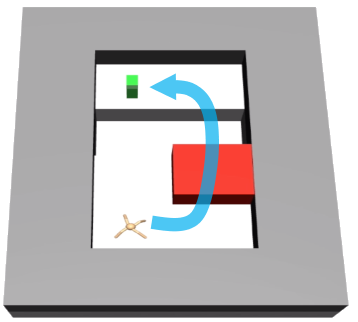}
        \end{subfigure}
        \begin{subfigure}{0.16\linewidth}
            \captionsetup{skip=0pt, position=below}
            \includegraphics[width=\linewidth]{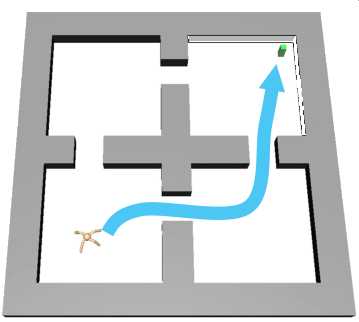}
        \end{subfigure}
        \begin{subfigure}{0.12\linewidth}
            \captionsetup{skip=0pt, position=below}
            \includegraphics[width=\linewidth]{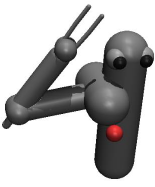}
        \end{subfigure}
        \begin{subfigure}{0.18\linewidth}
            \captionsetup{skip=0pt, position=below}
            \includegraphics[width=\linewidth]{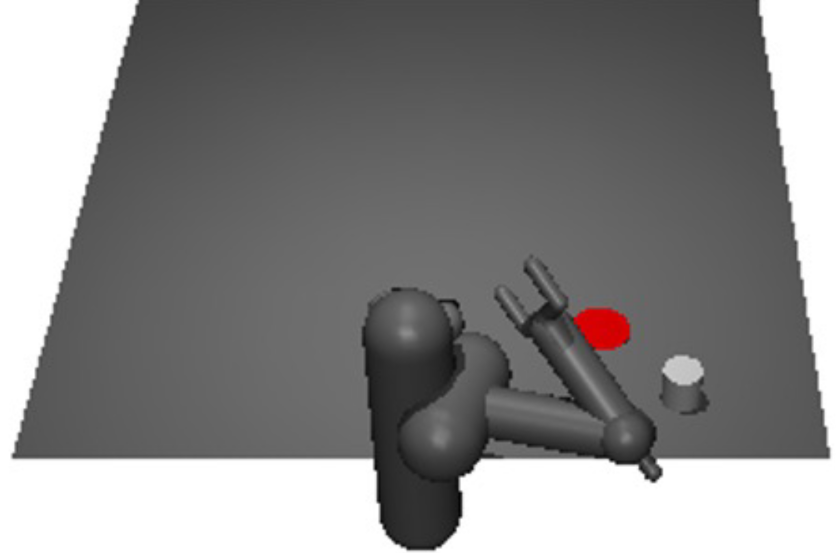}
        \end{subfigure}
        \caption{Environments used in our experiments.}
        \label{fig:envs}
    \end{figure*}


        \subsection{Efficient Online Subgoal Generation}
        
        During learning, we proposed a batch solution for HRL with latent GP subgoals that considers all the interconnected states in the trajectory. However, the inference involves matrix inversion of the covariance matrix $\mathbf{C}$ which grows with the number of states in the trajectory. Consequently, the inference complexity scales cubically with the number of states in the trajectory. During online HRL inference, the subgoal representation corresponding to states in the low-level trajectory follows a natural ordering, and thus our model can be relaxed to a direct graph, \ie, Markov chain. This formulation can be solved exactly without approximations by state-space form GP \citep{sarkka2012infinite,sarkka2013spatiotemporal} with a constant memory and computational complexity per state. 
        
        Specifically, the GP prior for latent subgoals can be transformed into a dynamical model for state-space GP inference, based on the hyperparameters $\gamma^{2}$, $\ell$ and $\sigma^2$ learned from training. The initial latent subgoal representation is estimated corresponding to Mat\'ern covariance function, \ie,  $\mathbf{z}_0 \sim \mathcal{N}(\boldsymbol{\mu}_0, \boldsymbol{\Sigma}_0)$ where 
        $\boldsymbol{\mu_0} = \mathbf{0}$ and $\boldsymbol{\Sigma_0} = \text{diag} \left(\gamma^2,   3\gamma^2/\ell    \right)$. As derived in  \citet{sarkka2013spatiotemporal}, an evolution operator which has the behavior of the Mat\'ern kernel is defined:
        \begin{equation}
        \mathbf{\Psi}_i=\exp \left[\left(\begin{array}{cc}
        0 & 1 \\
        -3 / \ell^{2} & -2 \sqrt{3} / \ell
        \end{array}\right) \Delta S_{i}\right],
        \end{equation}
        where the state difference $\Delta S_{i} = D(\mathbf{s}_i, \mathbf{s}_{i-1})$ is the distance between consecutive states. Then the subgoal representation is predicted by $\mathbf{z}_i | \mathbf{f}_{1:i-1}  \sim \mathcal{N}(\tilde{\boldsymbol{\mu}}_i, \tilde{\boldsymbol{\Sigma}}_i) $, where the mean and covariance are propagated as:
        \begin{align} 
        \tilde{\boldsymbol{\mu}}_i = \mathbf{\Psi}_i \boldsymbol{\mu}_{i-1},  \\
        \tilde{\boldsymbol{\Sigma}}_i = \mathbf{\Psi}_i \boldsymbol{\Sigma}_{i-1}\mathbf{\Psi}_i ^{\top} + \boldsymbol{\Omega}_i, 
        \end{align}
        where $\boldsymbol{\Omega}_i = \boldsymbol{\Sigma}_0 - \mathbf{\Psi}_i \boldsymbol{\Sigma}_0 \mathbf{\Psi}_i^{\top}$. The posterior mean and covariance is conditioned on the current intermediate latent representation $\mathbf{f}_i$:
        \begin{align} 
        \boldsymbol{\mu}_i = \tilde{\boldsymbol{\mu}}_i  + \mathbf{k}_i (f_i^{\top} - \mathbf{h}^{\top} \tilde{\boldsymbol{\mu}}_i ),\\
        \boldsymbol{\Sigma}_i = \tilde{\boldsymbol{\Sigma}}_i - \mathbf{k}_i \mathbf{h}^{\top} \tilde{\boldsymbol{\Sigma}}_i, 
        \end{align}
        where $\mathbf{k}_{i}=\tilde{\boldsymbol{\Sigma}}_i \mathbf{h} /\left(\mathbf{h}^{\top} \tilde{\boldsymbol{\Sigma}}_i  \mathbf{h}+\sigma^{2}\right)$ and the observation model $\mathbf{h}=\left(1 ~ 0\right)^{\top}$. The derivation of the above recursive update for the posterior mean and covariance for a new state $\mathbf{s}_i$ can be found in the Appendix. We note its resemblance to the Kalman Filter updates. 
        
        Note that the posterior latent subgoal representation $\mathbf{z}_i | \mathbf{f}_{1:i}  \sim \mathcal{N}(\boldsymbol{\mu}_i, \boldsymbol{\Sigma}_i) $ is conditioned on all state till the current time step and thus is able to encode longer-term memory of high-level actions.

    \begin{figure*}[!t]
        \centering
        \begin{subfigure}{0.24\linewidth}
            \captionsetup{skip=0pt, position=below}
            \includegraphics[width=\linewidth]{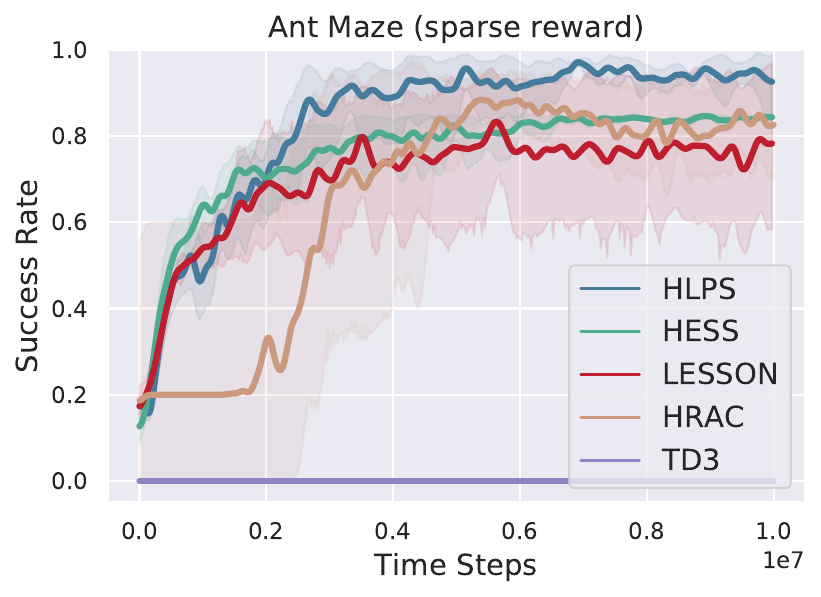}
        \end{subfigure}
        \begin{subfigure}{0.24\linewidth}
            \captionsetup{skip=0pt, position=below}
            \includegraphics[width=\linewidth]{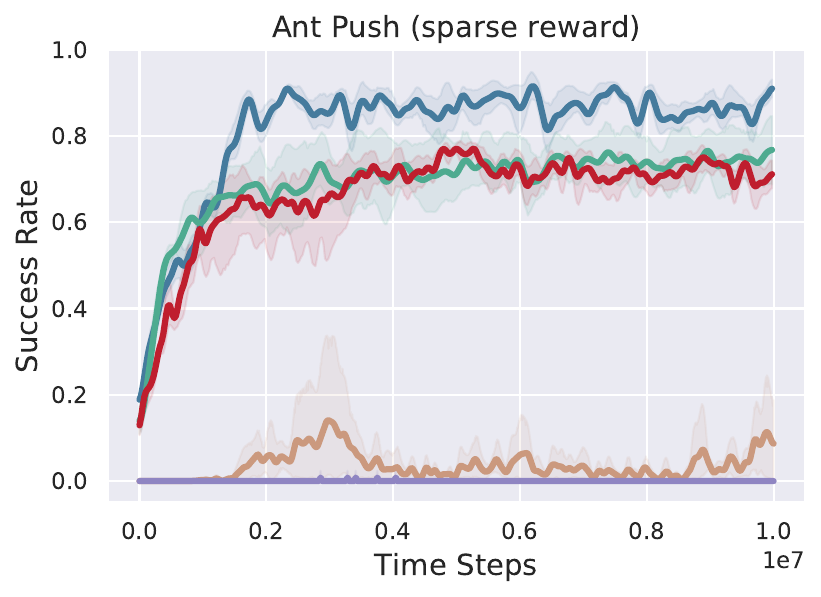}
        \end{subfigure}
        \begin{subfigure}{0.24\linewidth}
            \captionsetup{skip=0pt, position=below}
            \includegraphics[width=\linewidth]{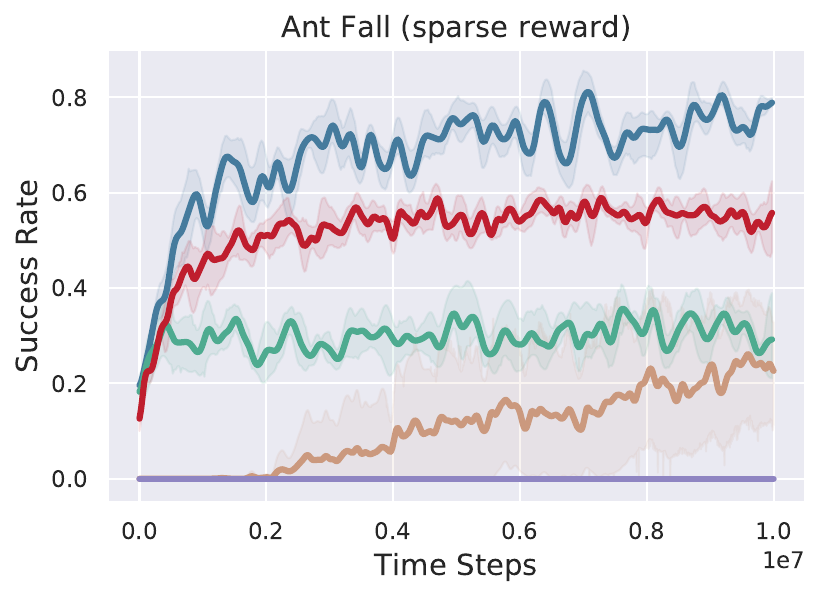}
        \end{subfigure}
        \begin{subfigure}{0.24\linewidth}
        \captionsetup{skip=0pt, position=below} 
            \includegraphics[width=\linewidth]{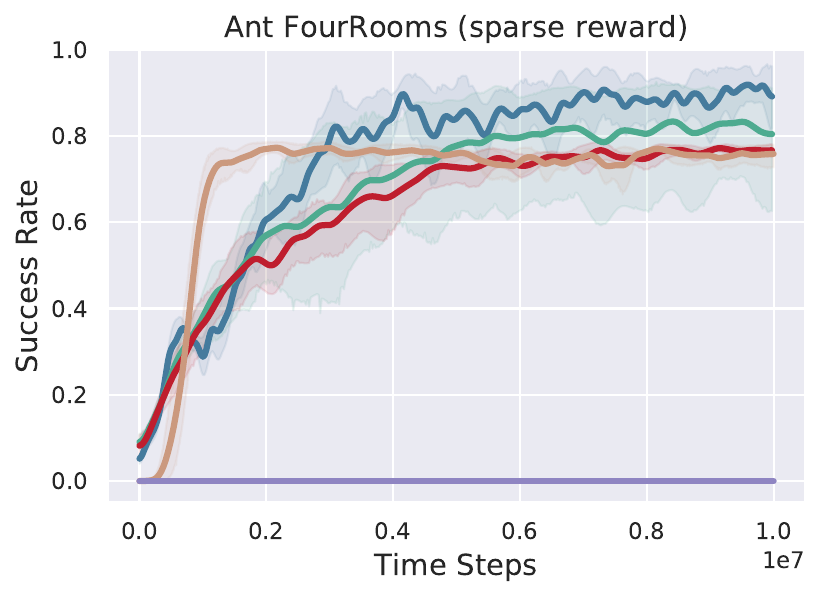}
        \end{subfigure}\\[-0.3ex] 
        \begin{subfigure}{0.24\linewidth}
            \captionsetup{skip=0pt, position=below} 
            \includegraphics[width=\linewidth]{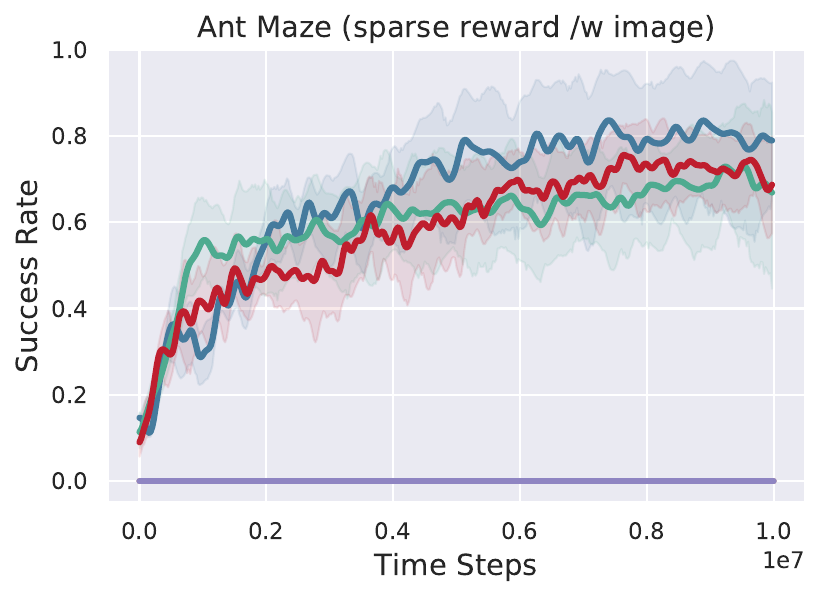}
        \end{subfigure}
        \begin{subfigure}{0.24\linewidth}
            \captionsetup{skip=0pt, position=below} 
            \includegraphics[width=\linewidth]{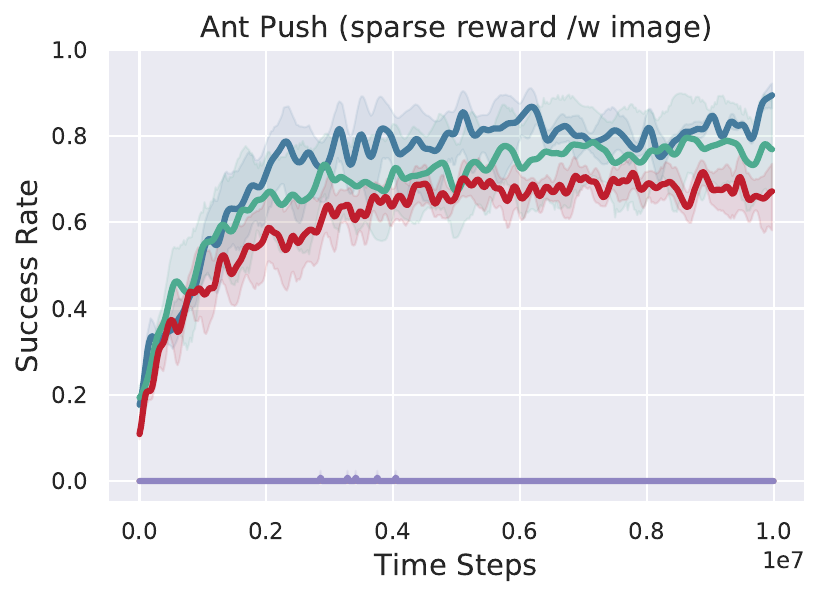}
        \end{subfigure}
        \begin{subfigure}{0.24\linewidth}
        \captionsetup{skip=0pt, position=below} 
            \includegraphics[width=\linewidth]{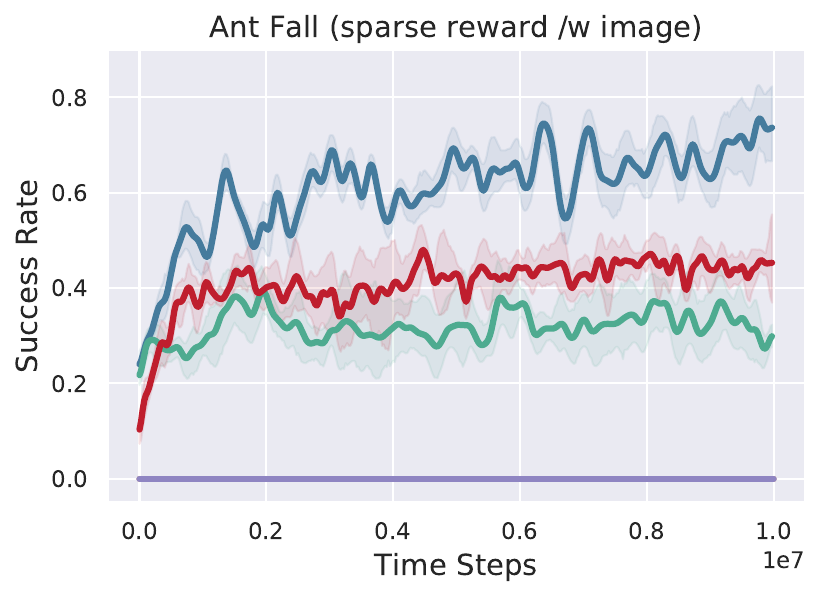}
        \end{subfigure}
        \begin{subfigure}{0.24\linewidth}
            \captionsetup{skip=0pt, position=below} 
            \includegraphics[width=\linewidth]{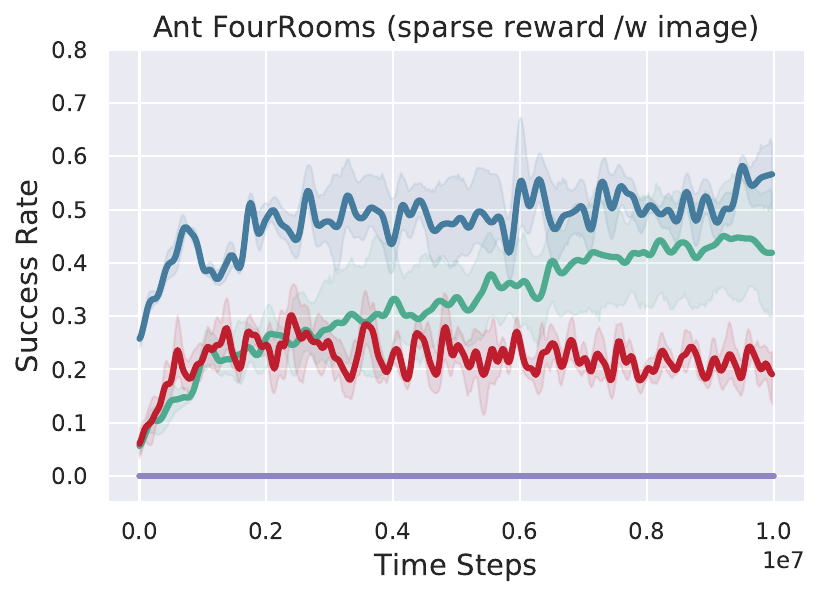}
        \end{subfigure}\\[-0.3ex] 
        \begin{subfigure}{0.24\linewidth}
            \captionsetup{skip=0pt, position=below} 
            \includegraphics[width=\linewidth]{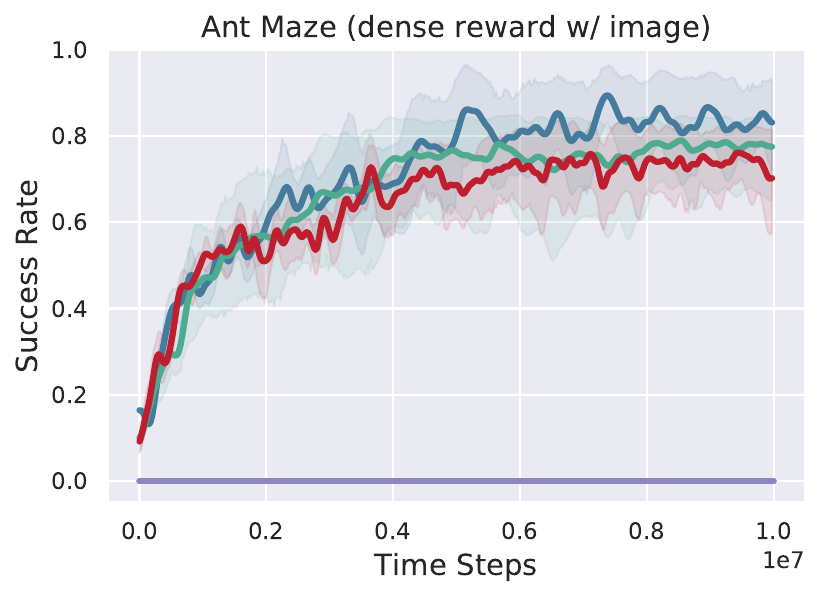}
        \end{subfigure}
        \begin{subfigure}{0.24\linewidth}
        \captionsetup{skip=0pt, position=below} 
            \includegraphics[width=\linewidth]{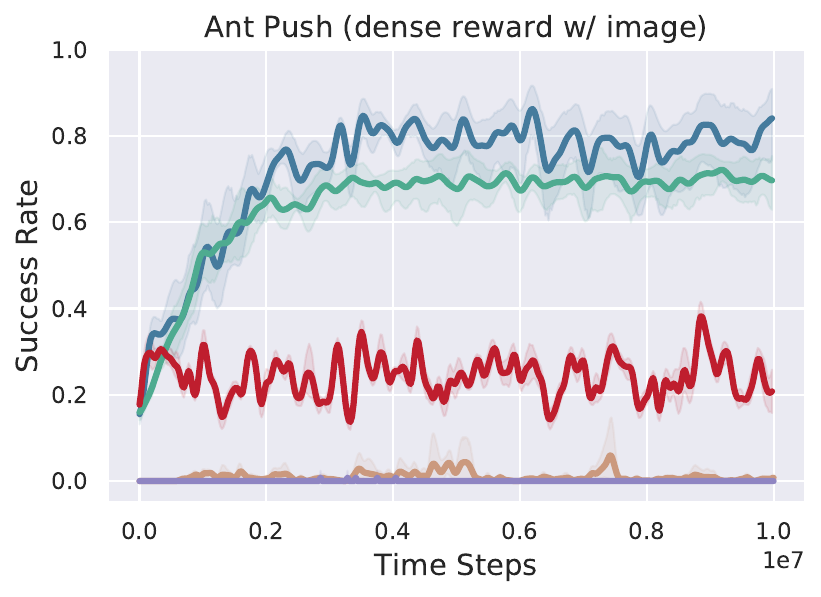}
        \end{subfigure}
        \begin{subfigure}{0.24\linewidth}
        \captionsetup{skip=0pt, position=below} 
            \includegraphics[width=\linewidth]{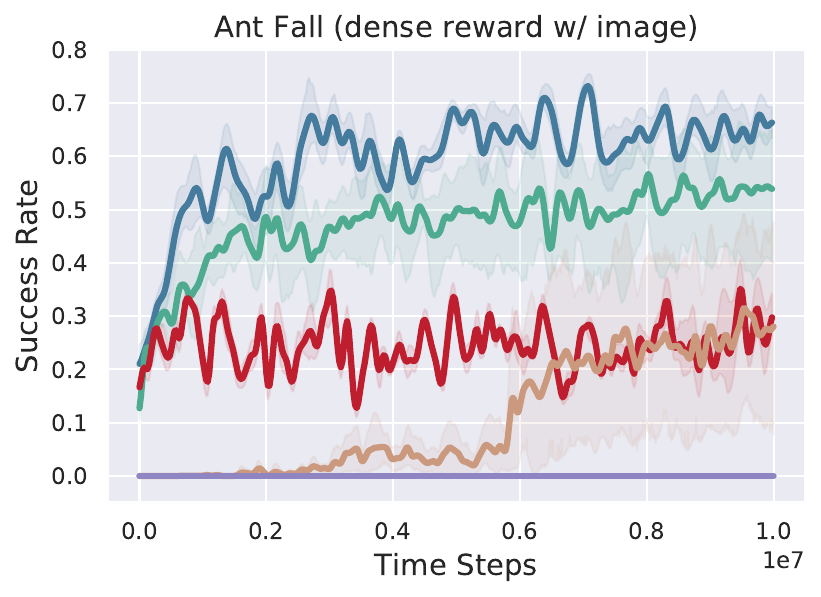}
        \end{subfigure}
        \begin{subfigure}{0.24\linewidth}
        \captionsetup{skip=0pt, position=below} 
            \includegraphics[width=\linewidth]{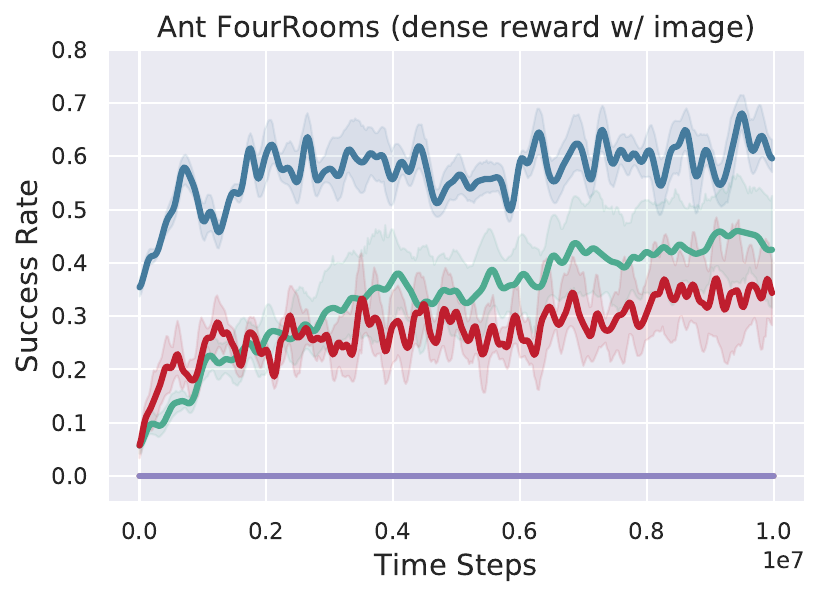}
        \end{subfigure}   
        \caption{Learning curves of our method and baselines in \textbf{stochastic} environments, with sparse (rows 1 and 2) or dense (row 3) external rewards, and with (rows 2 and 3) or without top-down image observations. Each curve and its shaded region represent the average success rate and 95\% confidence interval respectively, averaged over 10 independent trials.}
        \label{fig:comparison}
    \end{figure*}
    
    \begin{figure}[!t]
        \centering
        \begin{subfigure}{0.49\linewidth}
        \captionsetup{skip=0pt, position=below} 
            \includegraphics[width=\linewidth]{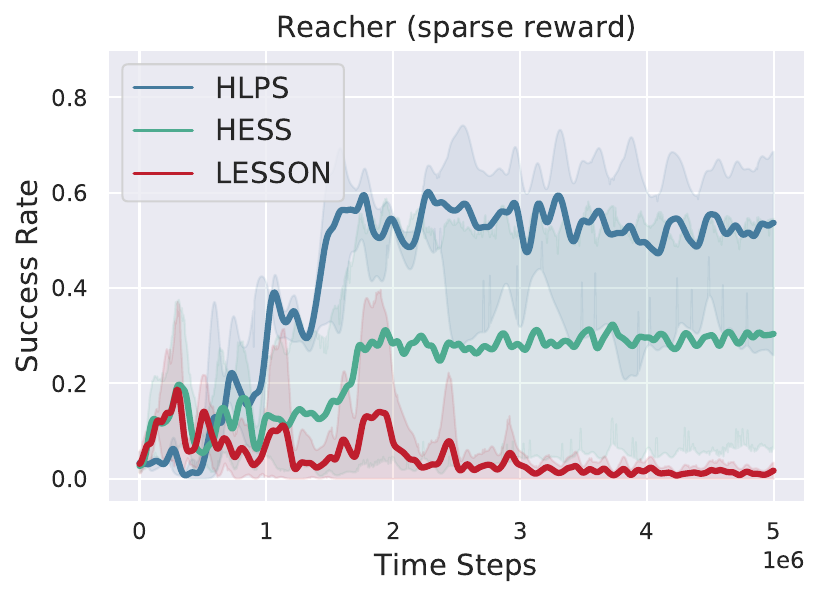}
            \caption{}
        \end{subfigure}
        \begin{subfigure}{0.49\linewidth}
        \captionsetup{skip=0pt, position=below} 
            \includegraphics[width=\linewidth]{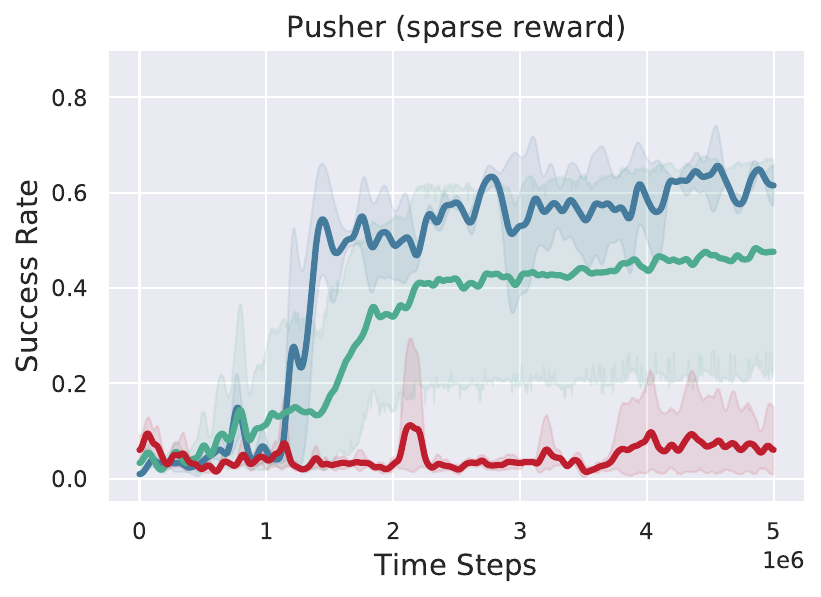}
            \caption{}
        \end{subfigure}\\[-0.3ex] 
        \caption{Learning curves of our method and baselines in robotic arm environments \textit{7-DOF Reacher} and \textit{7-DOF Pusher}, with sparse external rewards.}
        \label{fig:fetch}
    \end{figure}

    \begin{figure*}[!t]
        \centering
        
        \begin{subfigure}{0.99\linewidth}
            \captionsetup{skip=0pt, position=below} 
            \includegraphics[width=\linewidth]{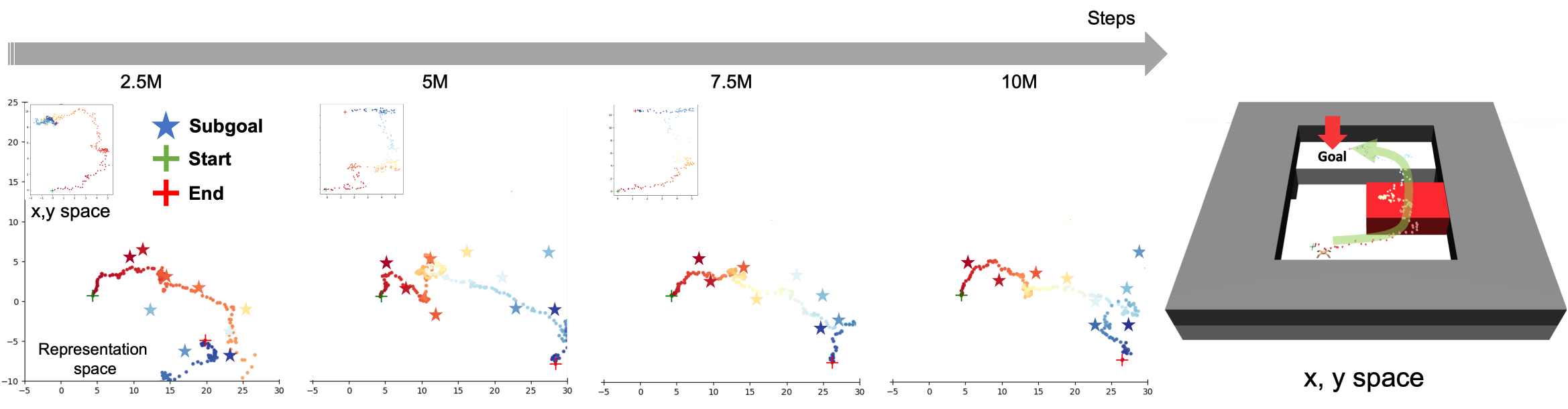}
            \caption{HLPS}
        \end{subfigure}\\[-0.3ex] 
        
        \begin{subfigure}{0.99\linewidth}
            \captionsetup{skip=0pt, position=below} 
            \includegraphics[width=\linewidth]{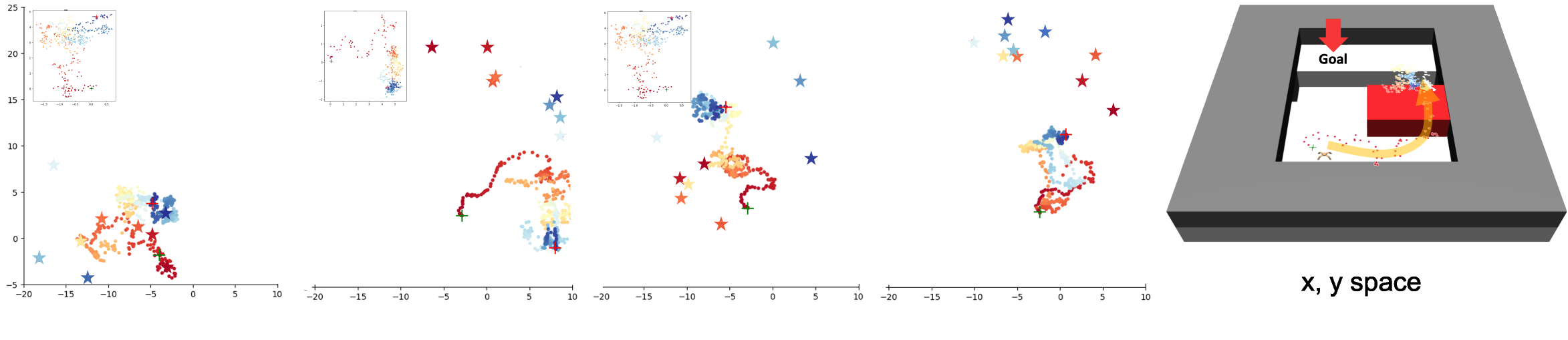}
            \caption{HESS}
        \end{subfigure}\\[-0.3ex] 
        
        \begin{subfigure}{0.99\linewidth}
            \captionsetup{skip=0pt, position=below} 
            \includegraphics[width=\linewidth]{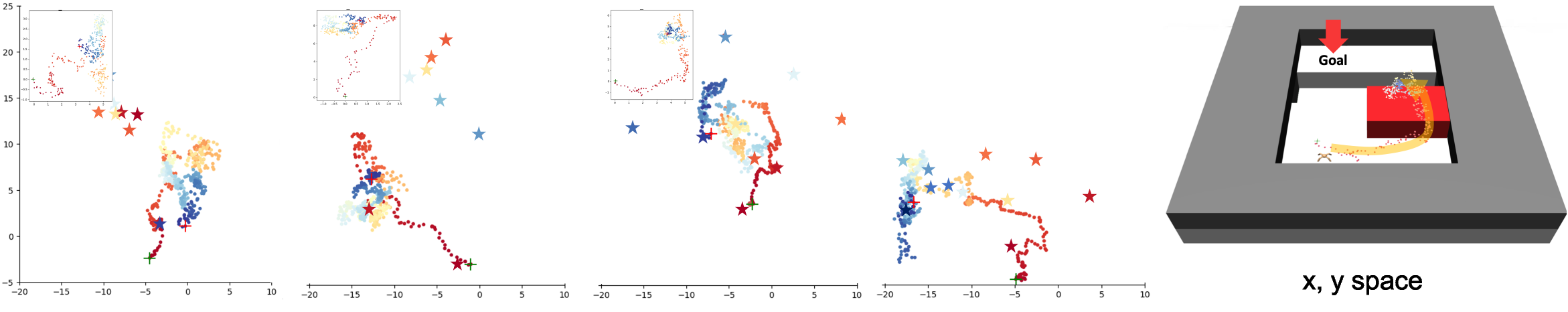}
            \caption{LESSON}
        \end{subfigure}
        
        \caption{Subgoal representation learning process in the challenging stochastic Ant Fall task with sparse reward. The illustration presents agent trajectories in the learned 2D latent subgoal space, transitioning from red to blue, and features color-coded stars marking subgoals. Additionally, a top-left minimap displays these trajectories within the x, y coordinate system. HLPS consistently learns stable subgoal representations over training, compared to HESS and LESSON - there is no significant change in the latent space from 5M steps until the end. The subgoals of HLPS align with low-level trajectories projected in the latent subgoal space, ensuring stable high-level transitions and low-level reward functions in unexplored state spaces. In contrast, HESS and LESSON exhibit poor subgoal reachability. HESS uses the counts across dramatically changing representation as novelty measure which misleads the exploration and generates unreachable subgoals. Both LESSON and HESS struggle to learn stable deterministic representations imposing local constraint in the presence of environmental stochasticity. In HLPS, distances in the latent subgoal space correlate with global transition counts, ensuring a representative distance between the start and goal of the maze. This global perspective helps to mitigate the local optima observed in HESS and LESSON, which arise from the local constraints applied during the training of deterministic subgoal representations. }
        \label{fig:visual}
    \end{figure*}
        
        \section{Related Work}
        
        Goal-conditioned HRL \citep{vezhnevets2017feudal,NachumGLL18,LevyKPS19,ZhangG0H020,wang2020i2hrl,li2020learning} where the high-level policy periodically generates subgoals to a low-level policy whilst the low-level policy learns how to efficiently reach these subgoals, has demonstrated great potentials in tackling temporally extended problems. A proper subgoal representation is crucial to goal-conditioned HRL since it defines the high-level action space and thus contributes to the stationarity of the high-level transition functions. Moreover, low-level behaviors can also be induced by dynamically changing subgoal space where the low-level reward function is defined. Hand-crafted space, \eg, predefining a subset of the state space as the subgoal space, is adopted in \citet{NachumGLL18,ZhangG0H020}. However, this approach requires domain knowledge and is limited to certain tasks. Using the whole state space has been investigated in \citet{LevyKPS19}, which is unscalable to tasks with high-dimensional observations. \citet{PereFSO18,NasirianyPLL19,NairF20} utilize variational autoencoder (VAE) \citep{KingmaW13} to compress high-dimensional observations in an unsupervised way, which, however, is unable to encode the states of hierarchical temporal scales in HRL. \citet{vezhnevets2017feudal} and \citet{DilokthanakulKP19} develop implicit subgoal representations by learning in end-to-end manner jointly with hierarchical policies. \citet{Sukhbaatar2018} develops a pre-training approach to learning subgoal representations via self-play. \citet{NachumGLL19} introduces the NOR approach by learning subgoal representations bounding the sub-optimality of hierarchical policies. \citet{li2020learning} develops a slowness objective for learning a deterministic subgoal representation function. Nevertheless, the existing methods have only proposed deterministic subgoal representations which may hinder effective explorations. Adopting the deterministic subgoal representation of \citet{li2020learning}, \citet{LiZWYZ22} develops an active exploration strategy to enhance the high-level exploration, by designing measures of novelty and potential for subgoals.

        Gaussian processes, which encode flexible priors over functions, are a probabilistic machine learning paradigm \citep{williams2006gaussian}. GPs have been used in other latent variable modeling tasks in RL. In \citet{engel2003}, the use of GPs for solving the RL problem of value estimation was first introduced. Then \citet{kuss2003gaussian} uses GPs to model the system dynamics and the value function. \citet{deisenroth2013gaussian} also develops a GP based transition model of a model-based learning system which explicitly incorporates model uncertainty into long-term planning and controller learning to reduce the effects of model errors. \citet{levine2011nonlinear} proposes  an algorithm for inverse reinforcement learning that represents nonlinear reward functions with GPs, which was able to recover both a reward function and the hyperparameters of a kernel function that describes the structure of the reward.

        \section{Experiments}	
        We evaluate our method in challenging environments
            with dense and sparse
            external rewards which require a combination of locomotion and
            object manipulation to demonstrate the effectiveness and
            transferability of our learned probabilistic subgoal
            representations. We compare our methods against standard RL
            and prior HRL methods. We also perform ablative studies to
            understand the importance of various components. \footnote{Code is available at https://github.com/vi2enne/HLPS}

        \subsection{Environments}
        We evaluate our approach on long-horizon continuous control tasks based on MuJoCo simulator \citep{todorov2012mujoco}, which are widely adopted in the HRL community. These tasks include \textit{Ant Maze}, \textit{Ant Push}, \textit{Ant Fall}, \textit{Ant FourRooms}, two robotic arm environments \textit{7-DOF Reacher} and \textit{7-DOF Pusher}  \citep{chua2018deep}, as well as four variants of Maze tasks featuring low-resolution image observations. 
     
     To evaluate the benefits of the proposed probabilistic subgoal representation, we make all Maze tasks more challenging in the following ways: (1) Environmental stochasticity: we enhance the robustness assessment of HLPS by introducing Gaussian noise with standard deviation $\sigma=0.1$ to the $(x, y)$ position of the agent at each step, following the precedent set by recent works such as HIGL \citep{kim2021landmark} and HRAC \citep{ZhangG0H020}. The results from deterministic environments are detailed in the Appendix.
        (2) Definition of ``success'': we tighten the success criterion to being within an $\ell^2$ distance of 1.5 from the goal, compared to a distance of 5 in \citet{NachumGLL18} and \citet{ZhangG0H020}.
         (3) External rewards: unlike the exclusive use of dense external rewards in \citet{NachumGLL18}, \citet{ZhangG0H020}, and \citet{li2020learning}, we also test settings with sparse external rewards, where a successful goal reach yields a reward of 1, and all other outcomes yield 0.
         (4) Random start/goal: Contrary to \citet{LiZWYZ22}, where the agent has fixed start and target positions, our tasks feature randomly selected start and target locations during training. For the \textit{7-DOF Reacher} task, we employ a 3D goal space representing the $(x, y, z)$ coordinates of the end-effector. In the \textit{7-DOF Pusher} task, the goal is to push an object to a 3D goal position. A success of an episode is defined to be if the goal is achieved at the final step of the episode. All methods undergo evaluation and comparison under the uniform task settings, ensuring a fair assessment \footnote{Further details, including environment specifics and parameter settings for experiment reproduction, are provided in the Appendix. }. 
    

     \begin{figure}[!t]
        \centering
            \begin{subfigure}[t]{0.49\linewidth}
                \centering
                \includegraphics[width=\linewidth]{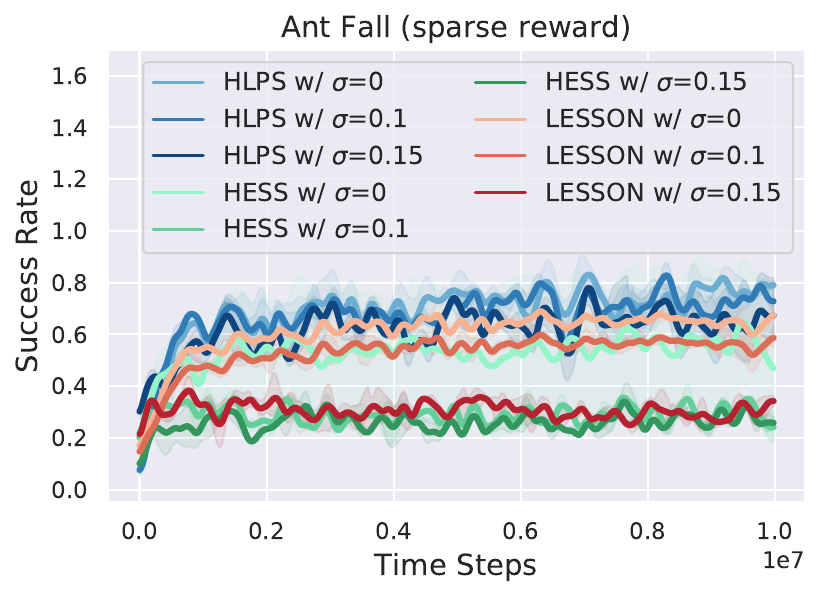}
            \end{subfigure}%
            \begin{subfigure}[t]{0.49\linewidth}
                \centering
                \includegraphics[width=\linewidth]{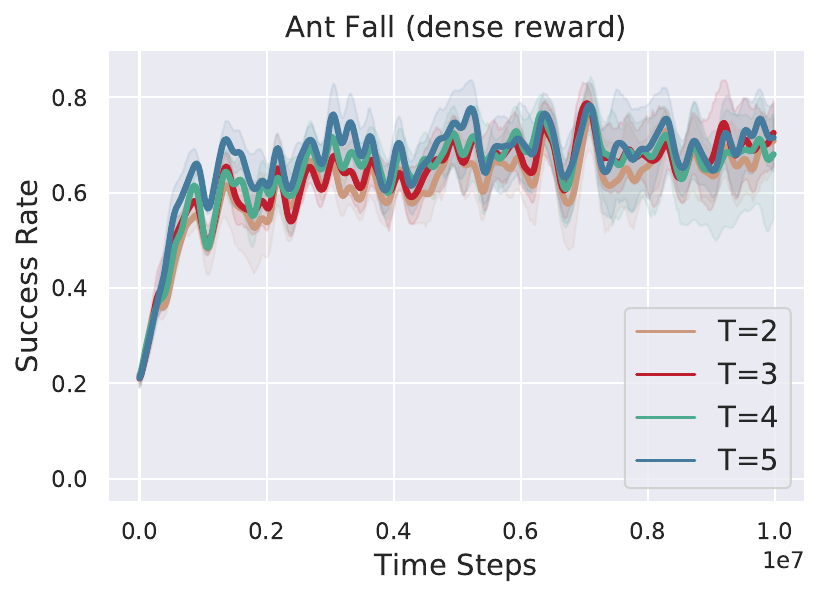}
            \end{subfigure}
            \caption{(Left) HLPS, HESS and LESSON on various levels of environmental stochasticities. (Right) HLPS for various time window sizes of the state set from Eq.~(\ref{eq:batch}), used in batch estimation of model hyperparameters.  }
            \label{fig:transfer}
    \end{figure}
    
    \begin{figure}[!t]
        \centering
            \begin{subfigure}[t]{0.49\linewidth}
                \centering
                \includegraphics[width=\linewidth]{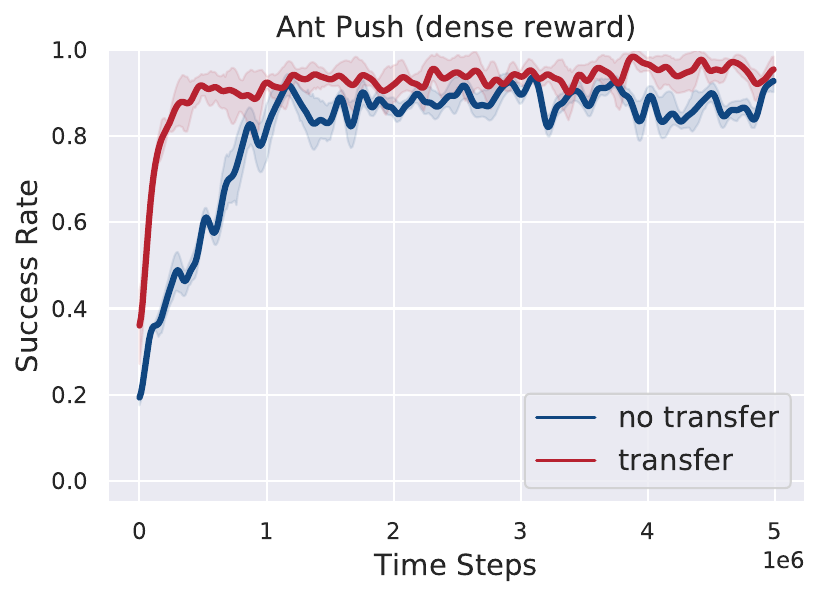}
            \end{subfigure}
            \begin{subfigure}[t]{0.49\linewidth}
                \centering
                \includegraphics[width=\linewidth]{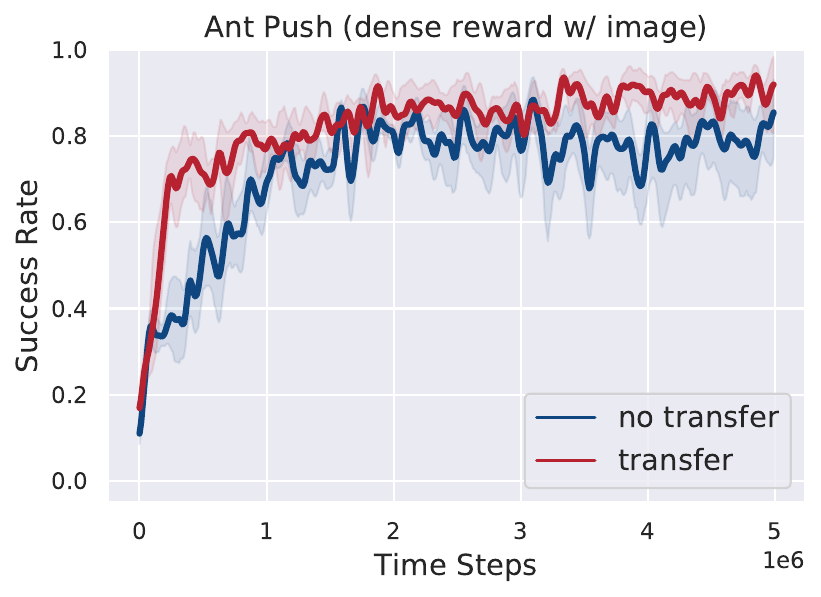}
            \end{subfigure}
            \caption{Transfer learning for the task (Left) \textit{Ant Fall} $\to$ \textit{Ant Push} and (Right) \textit{Ant Fall (Image)} $\to$ \textit{Ant Push (Image)}. The transferred subgoal representation and low-level policy enable superior sample efficiency and enhanced asymptotic performance.}
            \label{fig:transfer2}
    \end{figure}

    \begin{figure}[!t]
        \centering
        \begin{subfigure}{0.49\linewidth}
        \captionsetup{skip=0pt, position=below} 
            \includegraphics[width=\linewidth]{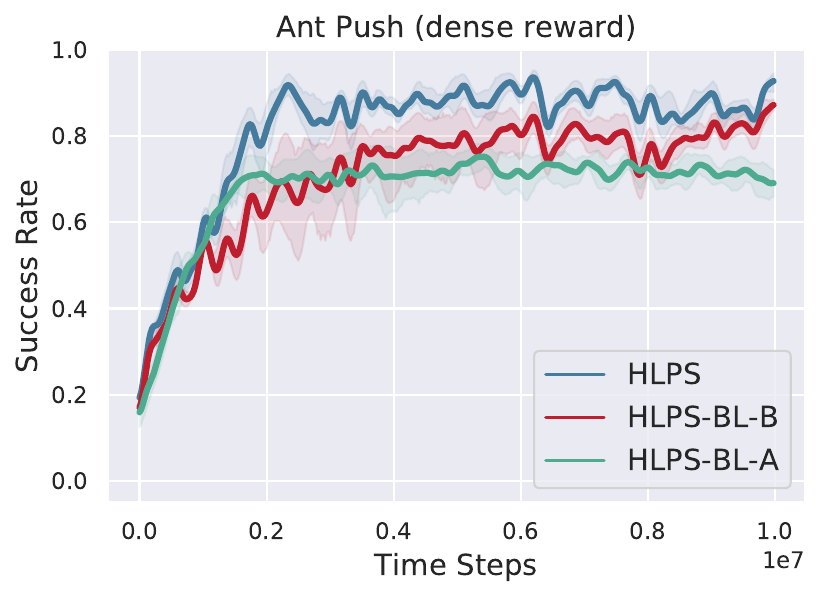}
            \caption{}
        \end{subfigure}
        \begin{subfigure}{0.49\linewidth}
        \captionsetup{skip=0pt, position=below} 
            \includegraphics[width=\linewidth]{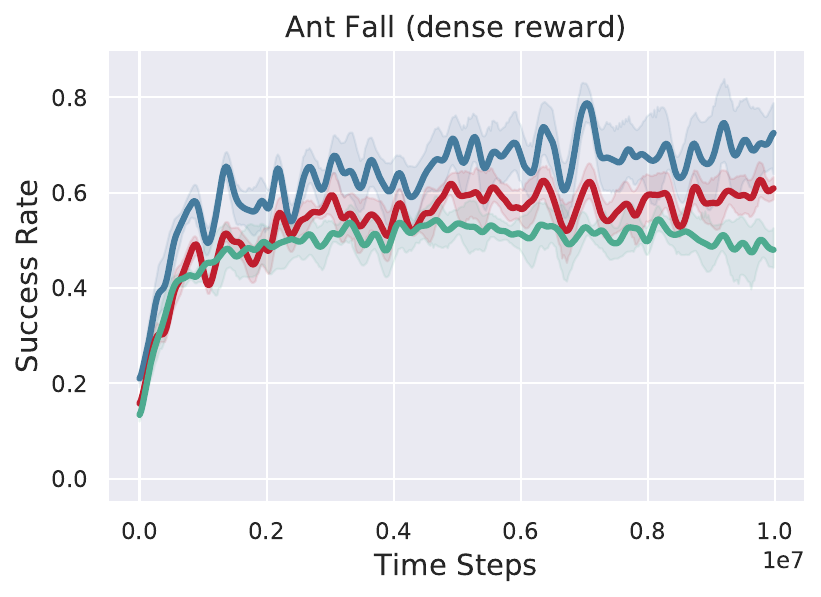}
            \caption{}
        \end{subfigure}\\[-0.3ex] 

        \caption{Ablation study comparing two baselines: HLPS-A omits our proposed learning objective and probabilistic subgoal representation, while HLPS-B enhances HLPS-A by incorporating the proposed probabilistic subgoal formulation and learning using an contrastive learning objective similar to that in \citet{li2020learning}. }
        \label{fig:ablation}
    \end{figure}

        \subsection{Analysis}
        We conduct experiments comparing to the following state-of-the-art baseline methods: (1) \textbf{LESSON} \citep{li2020learning}: a HRL algorithm that learns the deterministic subgoal representation. (2) \textbf{HESS} \citep{LiZWYZ22}: a HRL algorithm which introduces an active exploration strategy to LESSON \citep{li2020learning}. (3) \textbf{HRAC} \citep{ZhangG0H020}: a HRL algorithm which uses a pre-defined subgoal space. (4) \textbf{TD3}  \citep{fujimoto2018addressing}: a flat RL algorithm to validate the need for hierarchical policies. 
        
        \paragraph{Can HLPS surpass state-of-the-art HRL methods in learning stability, sample efficiency, and asymptotic performance?}
        Table \ref{table:quantitative} and Table \ref{table:quantitative_arm} in the Appendix show the final performance of the trained policy. Our method significantly outperforms all compared baselines. Fig.~\ref{fig:comparison} and Fig.~\ref{fig:fetch} illustrate the learning curves of our method and baselines across all tasks. Results for Maze tasks with dense external rewards provided in the Appendix. Our method out-performs all baselines in terms of stability, sample efficiency and asymptotic performance. The advantage of the probabilistic subgoal representation is more pronounced in the challenging \textit{Ant Fall} and \textit{Ant FourRooms} tasks, as well as the two more challenging robotic arm environments \textit{7-DOF Reacher} and \textit{7-DOF Pusher}. \textit{Ant Fall} requires both task and motion planning, while \textit{Ant FourRooms} uses a larger scale maze. Thus both tasks demand learning subgoal representation for unexplored areas. In the tasks with image input, the benefit of probabilistic subgoal representation of our method is more substantial, since learning the subgoal representation in a higher dimensional state space is more challenging and creates non-stationarities for deterministic subgoal representations in LESSON. The active exploration method introduced by HESS provides advantages in enhancing the generalization of deterministic subgoal representations in unexplored states (\eg, \textit{Ant FourRooms} with images) which is optimized for tasks with a fixed start and goal. However, its novelty measure, which combines counts in dynamically changing representation spaces, can potentially mislead exploration (as seen in \textit{Ant Fall}), especially when the goal is random. The results show a clear advantage of learned subgoal representations (HLPS, LESSON and HESS) compared to pre-defined (HRAC) subgoal spaces. Finally, the flat RL algorithm TD3 does not learn in the complex environments used in the experiments which further validates the need for hierarchical policies. In the \textit{7-DOF Reacher} and \textit{7-DOF Pusher} environments, similar comparisons can be made with HESS and LESSON, notably emphasizing LESSON's difficulties in effectively learning to achieve goals.

        \paragraph{Is HLPS capable of generating reachable subgoals to address the non-stationarity issue commonly encountered in off-policy training within HRL?}
         Fig.~\ref{fig:visual} illustrates the state embeddings learned at various training stages for the challenging \textit{Ant Fall} task with sparse external rewards. This allows for an intuitive comparison of subgoal representations acquired by HLPS, HESS, and LESSON. Notably, HLPS demonstrates stable evolution of subgoal representations throughout training, in contrast to HESS and LESSON. There are two in-depth observations: (1) The subgoals learned by HLPS are not only reachable but also largely align with the low-level trajectories projected into the latent subgoal space. This suggests that a stable subgoal representation enhances the stationarity of the high-level transitions and the low-level reward functions, providing strong learning signal even at the early stage of training. On the other hand, HESS and LESSON exhibit unstable embeddings, often leading to frequently shifting and distant subgoals. (2) In HLPS, the Euclidean distances in the latent space roughly correspond to the total number of transitions. More precisely, considering the number of transitions necessary for the agent to navigate between them, the start and goal positions in the maze should be distinctly separated in the latent space. However, due to the local constraints applied to the deterministic subgoal representations in both HESS and LESSON, the start and goal locations remain closely associated in the latent space. Consequently, many intermediate embeddings become stuck in local optima because they lack the global constraint present in HLPS. We underscore that our probabilistic representation learns the hyperparameters for the kernel function through finite number of support states, and then generalize to the entire space with a posterior distribution over the subgoal latent space.

    \paragraph{Does HLPS offer enhanced robustness in the face of environmental stochasticity?}
      We evaluate the robustness of HLPS against various environmental stochasticities and compare its performance with the deterministic subgoal representation approach LESSON, as well as with HESS. As illustrated in Fig.~\ref{fig:transfer} (Left), HLPS consistently outperforms both HESS and LESSON with increasing levels of Gaussian noise, specifically at \(\sigma\) values from the set \((0, 0.1, 0.15)\). Notably, HLPS demonstrates significantly smaller degradation in performance and lower variance in outcomes as environmental stochasticity increases, compared to the observed results in HESS and LESSON.
      
    \paragraph{Can HLPS effectively facilitate the transferability of learned subgoal space or low-level policies?}
        
        The generality of our GP based subgoal representation learning framework underpins transferable subgoal space as well as the low-level policy between different tasks of the same agent. To empirically experiment its transferability, the subgoal representation network, \ie, encoding layer and latent GP layer, and low-level policy network are initialized in a target task with the weights learned in a source task, with the rest of the network randomly initialized. Two pairs of source and target tasks, \ie,  \textit{Ant Fall} $\to$ \textit{Ant Push} and \textit{Ant Fall (Image)} $\to$ \textit{Ant Push (Image)}, are experimented. The learning curves on those two tasks are shown in Fig. \ref{fig:transfer2}, and we can observe that with the transferred subgoal representation and low-level policy the agent is more sample efficient and able to achieve higher performance. 
     
    \paragraph{How do various design choices within HLPS impact its empirical performance and effectiveness?}
    
    We conduct several ablation studies to analyze the design choices in our method. Initially, we compare our method, HLPS, with two baselines. HLPS-BL-A omits our proposed learning objective and probabilistic subgoal representation. In contrast, HLPS-BL-B builds upon HLPS-BL-A by incorporating the proposed probabilistic subgoal formulation and employing a contrastive learning objective akin to that used in \citet{li2020learning}.
      Fig. \ref{fig:ablation} shows the learning curves of various baselines. HLPS-BL-B exhibits much higher asymptotic performance than HLPS-BL-A but slightly lower performance than HLPS. This empirically demonstrates the effectiveness of our probabilistic subgoal representation and learning objective respectively. 
      
      We investigate the time window size of the set of states in Eq.~(\ref{eq:batch}) which are used to learn the model hyperparameters in batch estimation. As shown in Fig.~\ref{fig:transfer} (Right), increasing the time window size $T$ gives better performance at early training steps ($10^{6}\sim 7\times 10^{6}$) and eventually achieves similar performance as small time windows in larger training steps ($7\times 10^{6}\sim 10^{7}$). Our insight is that a larger time window gives more stable model hyperparameters with less training steps, which in turn induces sample-efficient stationarity of the policies due to stable subgoal representations. We report all other results based on time window $T=3$ without loss of generality.

        

        \section{Conclusion}
    
     This paper proposes a novel Gaussian process based method for learning probabilistic subgoal representations in Hierarchical Reinforcement Learning. Unlike existing approaches that focus on deterministic mappings, our model captures the posterior probability over the latent subgoal space.  This approach yields stability in previously unexplored state spaces, leading to stationarity in both the high-level transitions and low-level reward function. We also present a novel learning objective, integrating the learning of model hyperparameters and hierarchical policies within a unified framework. Our experiments demonstrate that this probabilistic subgoal representation significantly enhances sample efficiency, robustness against stochastic uncertainties, and asymptotic performance. Additionally, we show that our learned probabilistic subgoal representation facilitates the transfer of low-level policies between different tasks.
    
\section*{Acknowledgements}
We acknowledge CSC – IT Center for Science, Finland, for awarding this project access to the LUMI supercomputer, owned by the EuroHPC Joint Undertaking, hosted by CSC (Finland) and the LUMI consortium through CSC. We acknowledge the computational resources provided by the Aalto Science-IT project. V. Wang was funded by Research Council of Finland (353198). J. Pajarinen was partly funded by Research Council of Finland (353198).

\section*{Impact Statement}

This paper presents work whose goal is to advance the field of 
Machine Learning. There are many potential societal consequences 
of our work, none which we feel must be specifically highlighted here.


\bibliography{aaai22}
\bibliographystyle{icml2024}

\newpage
\appendix
\onecolumn
\section{Appendix}

\subsection{Algorithm}

 We provide Algorithm \ref{alg:gan} to show the training procedure of HLPS. Some details of subgoal latent space formulation $z$ are omitted for brevity, which refers to Eq. (\ref{eq:batch}). 



		\begin{algorithm}[!h]
			\caption{HLPS} 
			\label{alg:gan} 
			\begin{algorithmic}
				\STATE \textbf{Input:} High-level policy $\pi^{h}_{\theta_h}$, low-level policy $\pi^{l}_{\theta_l}$, encoding layer $f(\cdot)$, non-parametric latent GP layer with learnable hyperparameters ($\sigma^2$, $\gamma^{2}$ and $\ell$), GP update frequency $m$, higher-level action frequency $k$, number of training steps $N$, replay buffer $D$.
				\FOR {$n=1$ to $N$}
				\STATE Apply policies $\pi^{l}_{\theta_l}$ and $\pi^{h}_{\theta_h}$, collect experience $(s_{t}, g_{t}, a_t, r_t, s_{t+1}, g_{t+1})$\;
				\STATE Compute intrinsic reward $r_l (s_t, a_t, s_{t+1}, g_t) = - ||\phi(s_{t+1}) -g_t||_2$\;
				\STATE Update replay buffer $D$\;
				\STATE Update low-level policy $\pi^{l}_{\theta_l}$ and encoding layer $f(\cdot)$ with experience from replay buffer $D$ every timestep with Eq. (\ref{eq:objective})\;
				\STATE Update high-level policy $\pi^{h}_{\theta_h}$ with experience from replay buffer $D$ every $k$ timesteps\;
				\STATE Update latent layer hyperparameter with a batch of state transitions from replay buffer $D$ every $m$ timesteps with Eq. (\ref{eq:objective})\;
				\ENDFOR
			\end{algorithmic} 
		\end{algorithm}

	\subsection{Online Inference}
	
	The Gaussian process inference problem we formulated in the batch scheme can be rewritten in the form 
	\begin{equation}
	\begin{array}{l}
	\mathbf{z} \sim \mathcal{G P}\left(0, \kappa\left(\mathbf{s}, \mathbf{s'}\right)  \right) \\
	\mathbf{f}  = \mathbf{H} ~\mathbf{z(\mathbf{s})}  +\epsilon,  \epsilon \sim  \mathcal{N}(0, \sigma),
	\label{eq:pde}
	\end{array}
	\end{equation}
	where the linear operator $\mathbf{H}$ selects the training set inputs among the latent subgoal space values $\mathbf{H}~ \mathbf{z(\mathbf{s})}  = (\mathbf{z(\mathbf{s_1})},...,\mathbf{z(\mathbf{s_N})})$. This problem can be seen as an infinite-dimensional version of the Bayesian linear regression problem:
	\begin{equation}
	\begin{array}{l}
	\mathbf{z} \sim \mathcal{N}\left(0, K \right) \\
	\mathbf{f}  = \mathbf{H} ~\mathbf{z(\mathbf{s})}  +\epsilon
	\label{eq:pde2}
	\end{array}
	\end{equation}
	where $\mathbf{z}$ is a vector with Gaussian prior $\mathcal{N}\left(0, K \right) $ and $\mathbf{H} $ is constructed to select those elements of the vector $\mathbf{z}$ that can be actually observed \citep{sarkka2012infinite} . 
	
	This linear model can be extended such that the vector is allowed to change in time according to a linear stochastic differential equation (SDE) model and a new vector of measurements is obtained at times $t_i$ for $i = 1,...,T$ \citep{sarkka2019applied}:
	\begin{equation}
	\begin{array}{l}
	\frac{\partial\mathbf{z(t)} }{\partial t} =\mathbf{A}~\mathbf{z}(t) + \mathbf{L}  \mathbf{w}(t) \\
	\mathbf{f}_i  = \mathbf{H} ~ \mathbf{z}(t_i)  +\epsilon_i,
	\label{eq:kalman}
	\end{array}
	\end{equation}
	where $i=1,...T$,  $\mathbf{A}$, $\mathbf{L}$ and $\mathbf{H}$ are given matrices, $\epsilon_i$ is a vector of Gaussian measurements noises, and $\mathbf{w}(t) $ is a vector of white noise processes. The problem of estimating $\mathbf{z}(t)$ given the measurements can be solved using the classical Kalman filter and Rauch-Tung-Striebel (RTS) smoother. Assuming $\mathbf{z}(t_0) = \mathcal{N}(\boldsymbol{\mu}_0, \boldsymbol{\Sigma}_0)$,  evolution operator $\mathbf{\Psi}_i$, and $\boldsymbol{\Omega}_i = \boldsymbol{\Sigma}_0 - \mathbf{\Psi}_i \boldsymbol{\Sigma}_0 \mathbf{\Psi}_i^{\top}$, the filtering solution is recursively given by the following Kalman filter  \citep{sarkka2013spatiotemporal}: 
	
	- Prediction step:
	$$
	\begin{aligned}
	& \tilde{\boldsymbol{\mu}}_i=\mathbf{\Psi}_{i-1} \boldsymbol{\mu}_{i-1}, \\
	& \tilde{\boldsymbol{\Sigma}}_i=\mathbf{\Psi}_{i-1} \boldsymbol{\Sigma}_{i-1} \mathbf{\Psi}_{i-1}^{\top}+\boldsymbol{\Omega}_i
	\end{aligned}
	$$
	- Update step:
	$$
	\begin{aligned}
	\mathbf{v}_i & =\mathbf{y}_i-\mathbf{H}_i \tilde{\boldsymbol{\mu}_i}, \\
	\mathbf{S}_i & =\mathbf{H}_i \tilde{\boldsymbol{\Sigma}}_i \mathbf{H}_i^{\top}+\sigma^{2}, \\
	\mathbf{K}_i & =\tilde{\boldsymbol{\Sigma}}_i \mathbf{H}_i^{\top} \mathbf{S}_i^{-1} \\
	\boldsymbol{\mu}_i & =\tilde{\boldsymbol{\mu}}_i+\mathbf{K}_i \mathbf{v}_i, \\
	\boldsymbol{\Sigma}_i & =\tilde{\boldsymbol{\Sigma}}_i-\mathbf{K}_i \mathbf{S}_i \mathbf{K}_i^{\top}
	\end{aligned}
	$$
	
	The subgoal representation during \emph{online planning} can be formulated as spatio-temporal Gaussian process regression problem with models of the form
	\begin{equation}
	\begin{array}{l}
	\mathbf{z}(\mathbf{s},t) \sim \mathcal{G P}\left(0, \kappa\left(\mathbf{s}, t; \mathbf{s}, t^{'} \right)  \right)   \\
	\mathbf{f}_{i} = \mathbf{H}_i \mathbf{z}(\mathbf{s}, t_i) +\epsilon_i. 
	\label{eq:gp2}
	\end{array}
	\end{equation}
	By representing the temporal correlation as a stochastic differential equation kind of model and the spatial dimension as an additional vector element index, it is equivalent to the infinite-dimensional state space model \citep{sarkka2012infinite} as counterpart of model Eq. \ref{eq:kalman}:
	\begin{equation}
	\begin{array}{l}
	\frac{\partial\mathbf{z}(\mathbf{s},t) }{\partial t} = \mathcal{\mathbf{A}} ~\mathbf{z}(\mathbf{s},t)  + \mathbf{L}  \mathbf{w}(\mathbf{s},t) \\
	\mathbf{f_{i}}  = \mathbf{H}_i ~ \mathbf{z}(\mathbf{s}, t_i)  +\epsilon_i,
	\label{eq:pde}
	\end{array}
	\end{equation}
	where the latent state $\mathbf{z}(\mathbf{s},t) $ at time $t$ consists of the whole function  $\mathbf{s} \mapsto \mathbf{z}(\mathbf{s},t)$,  $\mathcal{\mathbf{A}}$ is 
	a $s\times s$ matrix of linear operators operating on $\mathbf{s}$, $\mathbf{L}\in \mathbb{R}^{s\times q}$,  $\mathbf{H}_i \in \mathbb{R}^{d\times s}$ are given matrices, $\mathbf{f}_{i} \in \mathbb{R}^d$, $\epsilon_i \sim  \mathcal{N}(0, \Sigma_i)$, and $\mathbf{w}(\mathbf{s},t)  \in \mathbb{R}^{q}$ is a Wiener process with a given diffusion matrix $\mathbf{Q}_c \in \mathbb{R}^{q\times q}$. 
	This formulation is an infinite-dimensional Markovian type of model, where the problem of estimating $\mathbf{z}(\mathbf{s},t) $ given the measurements can be similarly solved using the above Kalman filter resulting in the prediction and update steps in paper.
	
	\subsection{Environments}
	\label{env}
	
	\begin{enumerate}
		\item{} \textbf{Ant Maze}
		A `$\supset$'-shaped maze of size 12$\times$12 for a quadruped-Ant to solve a navigation task.  The ant needs to reach a goal position starting from a random position in a maze with dense rewards. It has  a continuous state space including the current position and velocity, the current time step t, and the goal location. During training, a random position is generated as the goal for each episode, and at each time step the agent receives a dense or sparse reward according to its negative Euclidean distance from the goal position. The success is defined as being within an Euclidean distance of 1.5 from the goal. At evaluation stage, the goal position is set to (0, 8). Each episode ends at 500 time steps. 
 		\item{} \textbf{Ant Push}: A challenging task that requires both task and motion planning. The agent needs to move to the left then move up and push the block to the right in order to reach the target. 
		\item{} \textbf{Ant Fall}: This task extends the navigation to three dimensions. The agent starts on a platform of height 4 with the target located across a chasm that it cannot cross by itself. The agent needs to push the block into the chasm and walk on top of it before navigating to the target. 
            \item{} \textbf{Ant FourRooms}: This Task requires
        the agent to navigate from one room to another
        to reach the exogenous goal. In this task, a larger (18 $\times$ 18) maze structure is used.

        \item{} \textbf{7-DOF Reacher/Pusher}: In this setup, a 7-DOF robot arm is situated in front of a table. 
        In the 7-DOF Reacher task, success is defined as the gripper reaching a specified 3D goal position. In the 7-DOF Pusher task, the primary objective is to maneuver a cylinder into this target. Each trial introduces variability with randomized positions of the cylinder and different initial velocities for the arm, though the target's location is fixed. The encompassed state space includes: (1) Joint positions and velocities of the robot (14 dimensions);
        (2) Center of mass of the cylinder (3 dimensions);
        (3) Center of mass of the gripper (3 dimensions).
        Collectively, these elements form a 20-dimensional state space.  Both tasks have a horizon of 100 steps.

		\item{} \textbf{Variants}: Another Ant Maze of size $24\times24$ with the same definition of ``success''  is used  (labeled `Large'). A variant (labeled `Image') with low-resolution image observations for each of the above task is adopted;  the observation is formed by zeroing out the x, and y coordinates and appending a 5$\times$5$\times$3 top-down view of the environment, as described in \cite{NachumGLL19,li2020learning}. Another variant with environmental stochasticity is also adopted - Gaussian noise with standard deviation $\sigma=0.1$ to the $(x,y)$ position of the ant robot at every step is added. 
		
		
		
	\end{enumerate}
	
	\subsection{Implementation}
	
	\subsubsection{Training and Evaluation Parameters}
	
	\begin{itemize}
		\item Learning rate of latent GP $1e-5$
		\item Latent GP update frequency 100
            \item Batch GP scheme time window size 3
		\item Subgoal dimension of size 2
		\item Learning rate 0.0002 for actor/critic of both levels
		\item Interval of high-level actions $k=50$
		\item Target network smoothing coefficient 0.005
		\item Reward scaling 0.1 for both levels
		\item Discount factor $\gamma = 0.99$ for both levels
		\item Learning rate for encoding layer 0.0001 
		\item Hierarchical policies are evaluated every 25000 timesteps by averaging over 10 randomly seeded trials 
	\end{itemize}
	
	\subsubsection{Network Architectures}
	We employ a two-layer hierarchical policy network similar to \cite{LevyKPS19,li2020learning} which adopts SAC \citep{haarnoja2018soft} for each level in the HRL structure. 
	Specifically, we adopt two networks each comprising three fully-connected layers (hidden layer dimension 256) with ReLU nonlinearities as the actor and critic networks of both low-level and high-level SAC networks. The output
	of the actor networks of both levels is activated using the tanh function and scaled according to the size of the environments. The encoding layer $f(\cdot)$ is parameterized by an MLP with one hidden layer of dimension 100 using ReLU activations. Adam optimizer is used for all networks.

	\subsubsection{Hardware}
	All of the experiments were processed using a single GPU (Tesla V100) and 8 CPU cores (Intel Xeon Gold 6278C @ 2.60GHz) with 64 GB RAM.

\begin{table*}[t]
	\addtolength{\tabcolsep}{-1pt}
	\centering
	\begin{small}
		\begin{tabular}{l|l|c|c|c|c|c}
			\cline{3-7} 
			\multicolumn{1}{c}{\color{white}\rule[-1ex]{1pt}{3ex}} &  &  HLPS & HESS & LESSON  & HRAC & TD3  \\
			\hline
			           & Dense & \textbf{0.90$\pm$0.04}  & 0.86$\pm$0.01  & 0.81$\pm$0.04 & 0.76$\pm$0.06 & 0.00$\pm$0.00 \\
			Ant Maze & Sparse & \textbf{0.93$\pm$0.05}  & 0.84$\pm$0.01  & 0.77$\pm$0.10 & 0.83$\pm$0.06 & 0.00$\pm$0.00 \\ 
			          & Dense /w image & \textbf{0.83$\pm$0.06}   & 0.78$\pm$0.05 & 0.73$\pm$0.05 & 0.00$\pm$0.00 & 0.00$\pm$0.00 \\
                        & Sparse /w image & \textbf{0.79$\pm$0.07}   & 0.67$\pm$0.12 & 0.71$\pm$0.05 & 0.00$\pm$0.00 & 0.00$\pm$0.00 \\
                \hline
			          & Dense & \textbf{0.93$\pm$0.01} & 0.80$\pm$0.04 & 0.71$\pm$0.02 & 0.01$\pm$0.00 & 0.00$\pm$0.00 \\
			Ant Push & Sparse & \textbf{0.91$\pm$0.01}  & 0.77$\pm$0.05  & 0.71$\pm$0.02 & 0.08$\pm$0.03 & 0.00$\pm$0.00  \\           
			          & Dense /w image  & \textbf{0.84$\pm$0.05}   & 0.70$\pm$0.03 & 0.24$\pm$0.01 & 0.01$\pm$0.01 & 0.00$\pm$0.00 \\
                        & Sparse /w image  & \textbf{0.87$\pm$0.03}   & 0.73$\pm$0.06 & 0.67$\pm$0.03 & 0.00$\pm$0.00 & 0.00$\pm$0.00 \\
                \hline
			          & Dense & \textbf{0.69$\pm$0.03}   & 0.54$\pm$0.01 & 0.49$\pm$0.03 & 0.11$\pm$0.09 & 0.00$\pm$0.00 \\
			Ant Fall & Sparse & \textbf{0.79$\pm$0.01}  & 0.29$\pm$0.05  & 0.54$\pm$0.02 & 0.24$\pm$0.07 & 0.00$\pm$0.00  \\  
			          & Dense /w image  & \textbf{0.66$\pm$0.01}   & 0.54$\pm$0.07 & 0.19$\pm$0.02 & 0.28$\pm$0.10 & 0.00$\pm$0.00 \\
                        & Sparse /w image  & \textbf{0.74$\pm$0.04}   & 0.30$\pm$0.02 & 0.32$\pm$0.01 & 0.00$\pm$0.00 & 0.00$\pm$0.00 \\
                \hline
			          & Dense & \textbf{0.93$\pm$0.02}   & 0.80$\pm$0.01 & 0.76$\pm$0.03 & 0.65$\pm$0.03 & 0.00$\pm$0.00 \\
			  Ant FourRooms & Sparse & \textbf{0.89$\pm$0.04}  & 0.82$\pm$0.08  & 0.77$\pm$0.01 & 0.76$\pm$0.01 & 0.00$\pm$0.00  \\  
			          & Dense /w image  & \textbf{0.61$\pm$0.02}   & 0.42$\pm$0.06 & 0.34$\pm$0.04 & 0.00$\pm$0.00 & 0.00$\pm$0.00 \\
                        & Sparse /w image  & \textbf{0.57$\pm$0.03}   & 0.42$\pm$0.07 & 0.21$\pm$0.01 & 0.00$\pm$0.00 & 0.00$\pm$0.00 \\
			\hline
		\end{tabular}
	\end{small}
	\caption{Final performance of the policy obtained after 10M steps of training, averaged over 10 randomly seeded trials with standard error. Comparisons are to \textbf{HESS} \citep{LiZWYZ22}, \textbf{LESSON} \citep{li2020learning}, \textbf{HRAC} \citep{ZhangG0H020},  and ``flat'' RL TD3 \citep{fujimoto2018addressing}. We can observe the overall superior performance of our method in stochastic environments, with dense or sparse external rewards and with or without top-down image observations.}
	\label{table:quantitative}
\end{table*}

\begin{table*}[t]
    \centering
    \begin{small}
        \begin{tabular}{l|c|c|c|}
            \hline
            & HLPS & HESS & LESSON \\
            \hline
            7-DOF Reacher & \textbf{0.62$\pm$0.03} & 0.48$\pm$0.13 & 0.06$\pm$0.04 \\ 
            \hline
            7-DOF Pusher & \textbf{0.54$\pm$0.11} & 0.21$\pm$0.10 & 0.01$\pm$0.01  \\   
            \hline
        \end{tabular}
    \end{small}
    \caption{Final performance of the policy obtained after 5M steps of training with sparse reward, averaged over 10 randomly seeded trials with standard error. Comparisons are to \textbf{HESS} \citep{LiZWYZ22} and \textbf{LESSON} \citep{li2020learning}.}
    \label{table:quantitative_arm}
\end{table*}

\begin{table}[t]
    \addtolength{\tabcolsep}{-1pt}
    \centering
    \begin{small}
        \begin{tabular}{l|c|c|c|c}
            \hline
            & Ant Maze & \specialcell {Ant Maze \\(Large)} & Ant Push   & Ant Fall    \\
            \hline
            HLPS           & \textbf{0.96$\pm$0.00 }  & \textbf{0.93$\pm$0.03 }   & \textbf{0.90$\pm$0.01} & \textbf{0.74$\pm$0.02}  \\
            \hline
            HESS            & 0.91$\pm$0.01 & 0.85$\pm$0.02   & 0.80$\pm$0.02 & 0.56$\pm$0.02  \\
            LESSON           & 0.89$\pm$0.06  & 0.74$\pm$0.15  & 0.74$\pm$0.02 & 0.54$\pm$0.03   \\
            HRAC            & 0.90$\pm$0.03 & 0.83$\pm$0.03   & 0.01$\pm$0.00 & 0.45$\pm$0.08  \\
            HIRO        & 0.71$\pm$0.02  & 0.57$\pm$0.05 & 0.00$\pm$0.00                        & 0.13$\pm$0.07             \\
            ORACLE        & 0.64$\pm$0.11  & 0.56$\pm$0.09  & 0.70$\pm$0.05                        & 0.28$\pm$0.09      \\
            TD3              & 0.00$\pm$0.00   & 0.00$\pm$0.00    & 0.01$\pm$0.00    & 0.00$\pm$0.00\\
            \hline
        \end{tabular}
    \end{small}
    \caption{Final performance of the policy obtained after 5M steps of training in deterministic environments, averaged over 10 randomly seeded trials with standard error. Comparisons are to \textbf{HESS} \citep{LiZWYZ22}, \textbf{LESSON} \citep{li2020learning}, \textbf{HRAC} \citep{ZhangG0H020}, \textbf{HIRO} \citep{NachumGLL18}, HRL with oracle subgoal space \textbf{Oracle}, and flat RL TD3 \citep{fujimoto2018addressing}. We can observe the overall superior performance of our method, which is consistent with the evaluation results in stochastic environments.}
    \label{table:quantitative2}
\end{table}

\begin{figure*}[!t]
    \centering
    
    \begin{subfigure}{0.48\linewidth}
    \captionsetup{skip=0pt, position=below} 
        \includegraphics[width=\linewidth]{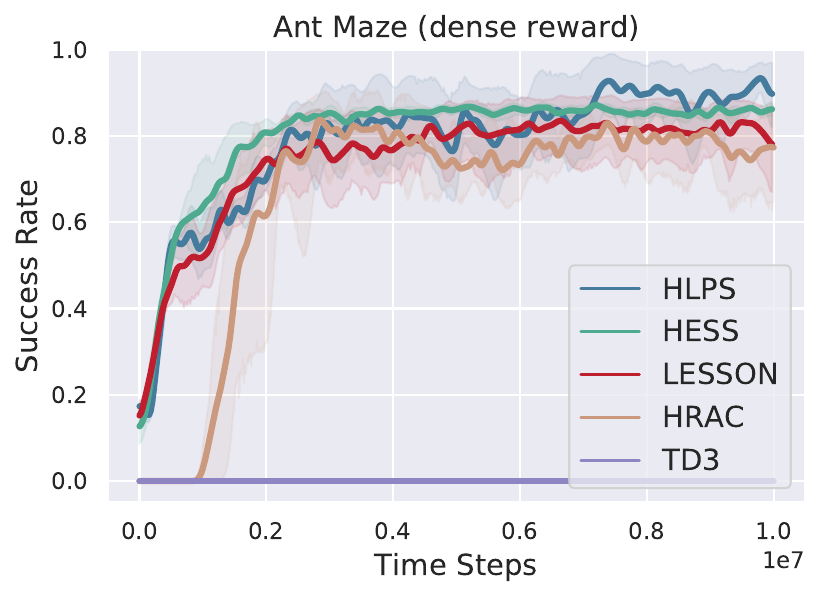}
    \end{subfigure}
    \begin{subfigure}{0.48\linewidth}
    \captionsetup{skip=0pt, position=below} 
        \includegraphics[width=\linewidth]{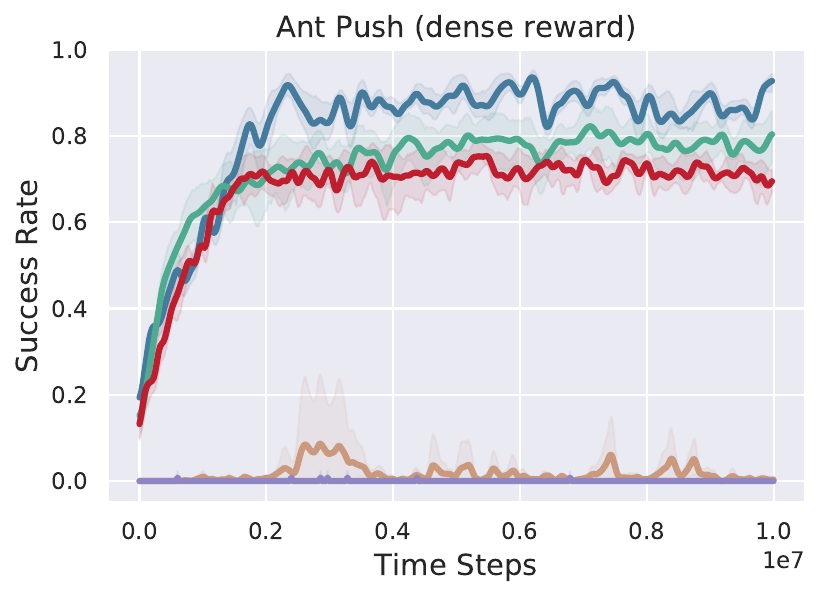}
    \end{subfigure}
    \begin{subfigure}{0.48\linewidth}
    \captionsetup{skip=0pt, position=below} 
        \includegraphics[width=\linewidth]{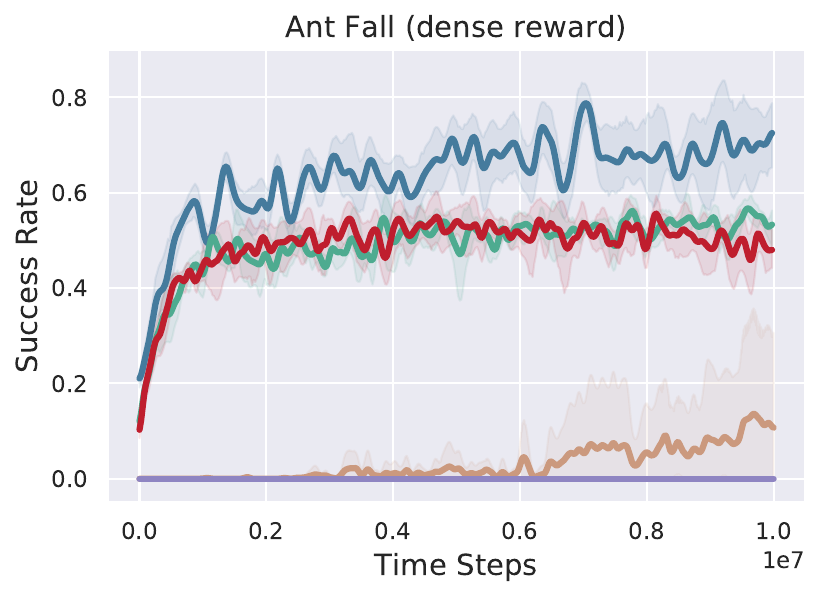}
    \end{subfigure}
    \begin{subfigure}{0.48\linewidth}
    \captionsetup{skip=0pt, position=below} 
        \includegraphics[width=\linewidth]{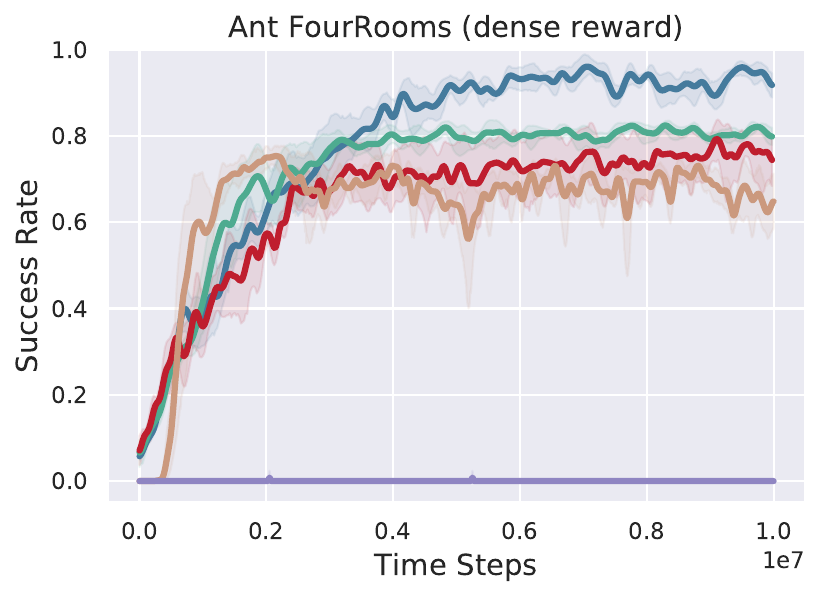}
    \end{subfigure}
    \caption{Learning curves of our method and baselines in \textbf{stochastic} environments with dense external rewards. }
    \label{fig:comparison2}
\end{figure*}
	
\begin{figure*}[!t]
    \centering
    
    \begin{subfigure}{0.48\linewidth}
    \captionsetup{skip=0pt, position=below} 
        \includegraphics[width=\linewidth]{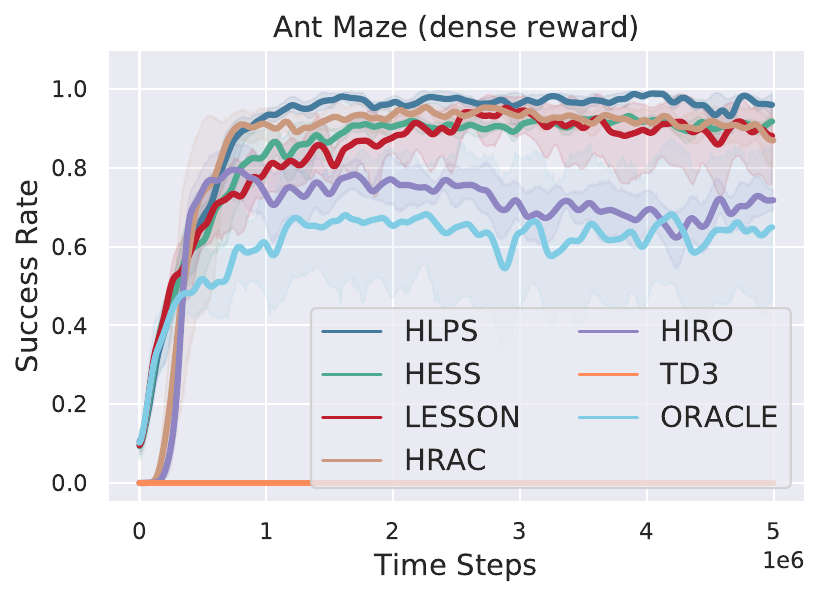}
        \caption{}
    \end{subfigure}
    \begin{subfigure}{0.48\linewidth}
    \captionsetup{skip=0pt, position=below} 
        \includegraphics[width=\linewidth]{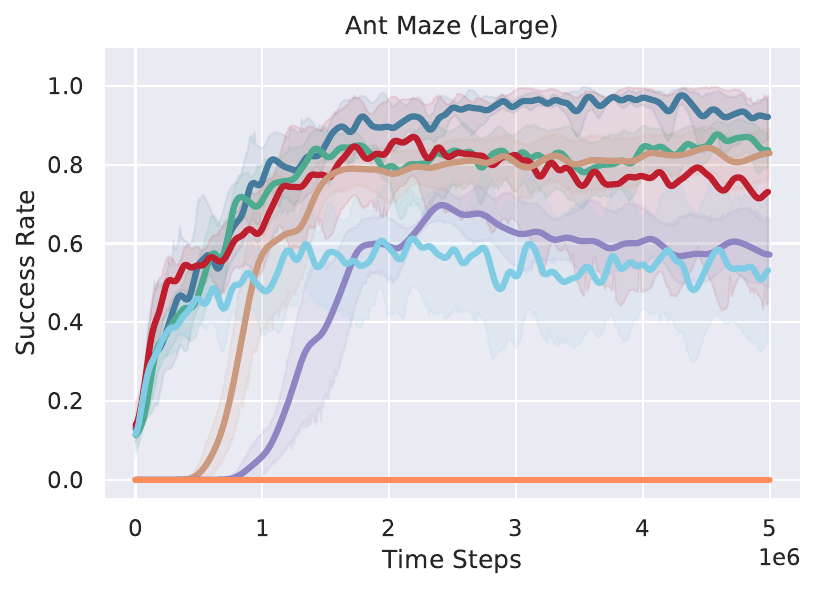}
        \caption{}
    \end{subfigure}
    \begin{subfigure}{0.48\linewidth}
    \captionsetup{skip=0pt, position=below} 
        \includegraphics[width=\linewidth]{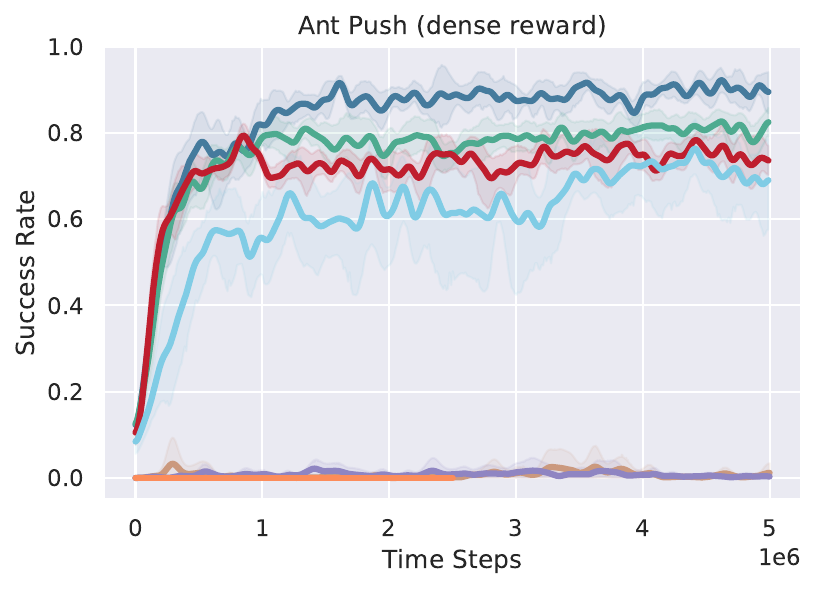}
        \caption{}
    \end{subfigure}
    \begin{subfigure}{0.48\linewidth}
    \captionsetup{skip=0pt, position=below} 
        \includegraphics[width=\linewidth]{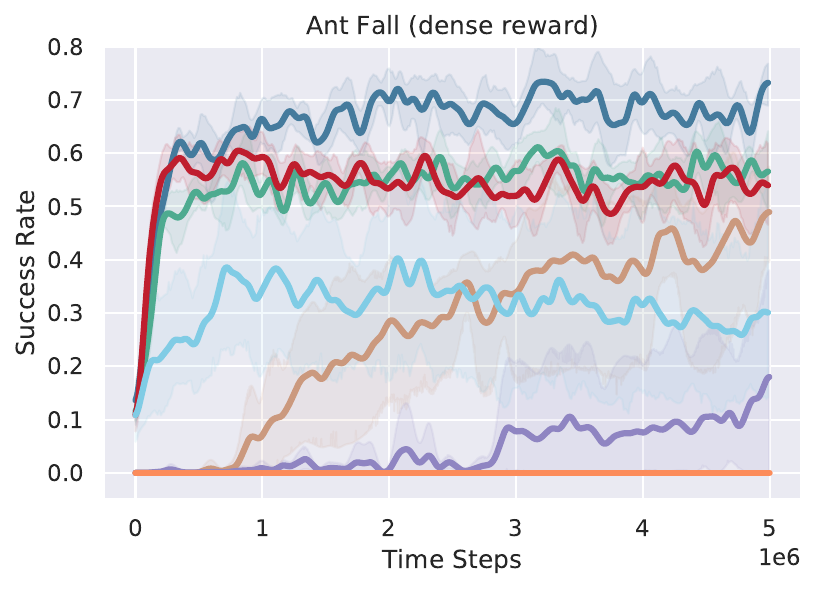}
        \caption{}
    \end{subfigure}\\[-0.3ex] 
    
    \caption{Learning curves of our method and baselines in \textbf{deterministic} environments with dense external rewards. }
    \label{fig:comparison3}
\end{figure*}
 \subsection{Additional Experiments}
 We show the learning curves of our method and baselines in stochastic environments with dense external rewards in Fig. \ref{fig:comparison2}, and its quantitative evaluation results can be found in Table \ref{table:quantitative}.

Additionally, we evaluate on deterministic Ant Maze, Ant Push and Ant Fall, as well as a `large' Ant Maze of size $24\times24$, with dense external reward. These experiments are conducted in comparison to \textbf{HESS} \citep{LiZWYZ22}, \textbf{LESSON} \citep{li2020learning}, \textbf{HRAC} \citep{ZhangG0H020} and \textbf{TD3}  \citep{fujimoto2018addressing}, as well as the following two baseline methods\footnote{We use the official implementations \url{https://github.com/SiyuanLee/LESSON}, \url{https://github.com/SiyuanLee/HESS/}, \url{https://github.com/trzhang0116/HRAC} and \url{https://github.com/sfujim/TD3}.}: 
	\begin{enumerate}
		\item\textbf{Oracle}: HRL with the oracle subgoal space, \ie, $x$, $y$ coordinates of the agent, in navigation tasks. 
		\item \textbf{HIRO} \citep{NachumGLL18}: an off-policy goal-conditioned HRL algorithm using a pre-defined subgoal space. 
	\end{enumerate} 
 Note, all methods are evaluated and compared using the same settings of tasks. Table \ref{table:quantitative2} shows the comparative results on deterministic environments, and Fig. \ref{fig:comparison3} shows the learning curves of all baselines.

\end{document}